\documentclass[runningheads]{llncs}
\usepackage[T1]{fontenc}
\usepackage{graphicx}
\usepackage{booktabs}
\usepackage[misc]{ifsym}
\newcommand{\corr}{(\Letter)}
\usepackage{mwe}
\usepackage{times}
\usepackage{soul}
\usepackage{url}

\usepackage[hidelinks]{hyperref}

\usepackage{amsmath}
\usepackage{amssymb}

\usepackage{dsfont}
\usepackage{booktabs}
\usepackage{algorithm}
\usepackage{algorithmic}
\usepackage[switch]{lineno}

\usepackage{comment}
\usepackage{mathtools}
\usepackage{xcolor}

\usepackage{subcaption}
\usepackage{multirow}
\usepackage{bm}

\newtheorem{prop}{Proposition}

\newcommand{\ie}{\textit{i}.\textit{e}.,}
\newcommand{\eg}{\textit{e}.\textit{g}.,}

\newcommand{\sota}{state-of-the-art}
\renewcommand{\eqref}[1]{Eq.\,(\ref{#1})}

\newcommand{\romsm}[1]{\lowercase\expandafter{\romannumeral #1\relax}}

\DeclarePairedDelimiter{\norm}{\lVert}{\rVert}

\begin{document}

\title{Towards Interpretable Adversarial Examples via Sparse Adversarial Attack}

% \titlerunning{Sparse Adversarial Attack}
% If the full title of your paper is short enough to also fit in the running head, you can omit the abbreviated paper title here. You can check as follows: if you comment out the \titlerunning line, something will appear in the header of all odd-numbered pages of your PDF from page 3 onward. This something is either the full title (in which case all is well), or the error message "Title Suppressed Due to Excessive Length". If this error message appears, you're going to want to provide an abbreviated title within the \titlerunning command, because if you won't do it, Springer will do it for you.

%N.B.: Author information (both in the \author{} and \authorrunning{} command) should only be present in the Camera-Ready Version of your paper. The version that you initially submit for review, ought to be double-blind. So, when initially submitting your paper, use:
% \author{Author information scrubbed for double-blind reviewing}
%
\author{Fudong Lin\inst{1} \and
Jiadong Lou\inst{1}  \and
Hao Wang\inst{2} \and
Brian Jalaian\inst{3} \and
% Antwan Andr\'e Patton\inst{1}\orcidID{0000-1111-2222-3333} \and
Xu Yuan\inst{1} \corr}

% You may leave out the orcidID information, if you want to.
% Use \corr to indicate the corresponding author. Note the spacing around the \corr command. Only one author can be the corresponding author.

%N.B.: comment out the \authorrunning{} command for the double-blind version of your paper submitted for review. Later, if your paper is accepted, use the command for the Camera-Ready Version.
\authorrunning{F. Lin, J. Lou, et al.}
% If there are two authors, write 'A.L. Benjamin and C.C. Broadus Jr.'
% If there are more than two authors, '[...] et al.' is used.

\institute{University of Delaware, Newark, DE 19716, USA \\
\email{\{fudong,loujd,xyuan\}@udel.edu} \and
Stevens Institute of Technology, Hoboken, NJ 07030, USA \\
\email{hwang9@stevens.edu}
\and University of West Florida, Pensacola, FL 32514, USA
\email{brian.a.jalaian.civ@mail.mil}
}

\maketitle              % typeset the header of the contribution

\begin{abstract}
    Sparse attacks are to optimize the magnitude of adversarial perturbations 
    for fooling deep neural networks (DNNs) involving only a few perturbed pixels (\ie\ under the $l_{0}$ constraint),
    suitable for interpreting the vulnerability of DNNs.
    However, existing solutions fail to yield interpretable adversarial examples due to their poor sparsity.
    Worse still, they often struggle with heavy computational overhead,
    poor transferability, and weak attack strength.
    In this paper, we aim to develop a sparse attack
    for understanding the vulnerability of DNNs
    by minimizing the magnitude of initial perturbations 
    under the $l_{0}$ constraint, 
    to overcome the existing drawbacks while achieving a fast, transferable, and strong attack to DNNs.
    In particular, a novel and \textit{theoretical sound} parameterization technique is introduced to approximate the  NP-hard $l_{0}$ optimization problem, 
    making directly optimizing sparse perturbations computationally feasible.
    Besides, a novel loss function is designed to augment initial perturbations by maximizing the adversary property 
    and minimizing the number of perturbed pixels simultaneously.
    Extensive experiments are conducted to demonstrate that 
    our approach, with theoretical performance guarantees, outperforms state-of-the-art sparse attacks 
    in terms of computational overhead, 
    transferability, and attack strength, 
    expecting to serve as a benchmark for evaluating the robustness of DNNs. 
    In addition, theoretical and empirical results validate that 
    our approach yields sparser adversarial examples,
    empowering us to discover two categories of noises, \ie\ ``obscuring noise'' and ``leading noise'', 
    which will help interpret how adversarial perturbation misleads the classifiers into incorrect predictions.
    Our code is available at \textcolor{magenta}{\url{https://github.com/fudong03/SparseAttack}}.
\keywords{Sparse Attack  \and Adversarial Attack \and Interpretability.}
\end{abstract}

\section{Introduction}
\label{sec:intro}
% \vspace{-0.5 em}

Deep neural networks (DNNs) have demonstrated impressive performance on a range of challenging tasks, including image classification~\cite{bartlett2012imagenet,simonyan2014very,he2016deep,dosovitskiy:iclr21:vit}, natural language processing~\cite{vaswani2017attention,devlin2019bert}, and various other domains~\cite{jumper:nature21:alpha_fold,yi:ijcai21:oversampling,fudong:iccv23:mmst_vit,rives:pans21:ESM-1b,fudong:ecml23:storm,ma2024malletrain,ali2025enabling,lan2025contextual,zawad2025fedcust,tiankuo:iclr25:bonemet}.
However, recent studies~\cite{szegedy2014intriguing,goodfellow2015explaining} have revealed a critical vulnerability: DNNs can be easily fooled by adversarial examples. These examples are generated by adding small, human-imperceptible perturbations to natural images, causing the models to make incorrect predictions with high confidence.
This vulnerability leads to severe security threats on DNNs,
\eg\  a prior study \cite{eykholt2018robust} reported that 
adding human-imperceptible 
perturbation to a {\em Stop} sign 
made \sota\ classifiers misclassify it as a Speed Limit $45$,
thereby hindering  DNNs' wide applicability to such security-critical domains as
face recognition~\cite{kurakin2017adversarial,thys2019fooling}, 
autonomous driving~\cite{eykholt2018robust,kong2020physgan}, \textit{etc}.

The mainstream attack strategy targeting DNNs is 
to optimize the magnitude of adversarial perturbations.
In general, a perturbation is constrained by $l_{p}$ norm, with $p=0, 1, 2,$ or $\infty$,
and can be categorized into two clusters, 
\ie\ dense attack and sparse attack.
The former needs to modify almost all pixels 
under the $l_{2}$ or $l_{\infty}$ constriant~\cite{dong2020greedyfool},
while the latter %aims to just 
perturbs a few pixels 
under the $l_{0}$ (or sometimes $l_{1}$) constriant~\cite{zhu2021homotopy}.

To date, Fast Gradient Sign Method (FGSM)-based
approaches~\cite{goodfellow2015explaining,kurakin2017adversarial,dong2018boosting,xie2019DI-FGSM,dong2019TI-FGSM}
are known to be prominent dense attacks, 
because they arrive at fast and highly transferable adversarial attacks 
by optimizing adversarial perturbations under the $l_{\infty}$ constraint.
However, they tend to perturb almost all pixels,
making them hard to interpret adversarial attacks 
due to their overly perturbed adversarial examples.
In sharp contrast, 
sparse attacks~\cite{su2019one,modas2019sparse,fan20factor,dong2020greedyfool,zhu2021homotopy,maura:nips21:fmn,croce:aaai22:sparse_rs,williams:cvpr23:sa_moo,ming:cvpr24:egs_tssa,vo:iclr24:brusle_attack} 
minimize the $l_{0}$ distance between natural images and adversarial examples,
for attacking DNNs with only a few perturbed pixels.
Hence, sparse attacks usually provide additional insights into adversarial attacks,
able to better interpret the vulnerability of DNNs~\cite{fan20factor}.
However, optimizing the magnitude of perturbations under the $l_{0}$ constraint falls into the NP-hard problem, so previous solutions often get trapped in local optima~\cite{modas2019sparse,zhu2021homotopy}, making the resulted attacks possess an insufficient adversary property. 
As such, existing sparse attacks suffer from the drawbacks of heavy computational overhead
\cite{dong2020greedyfool}, poor transferability, and weak attack intensity.
Worse still, their resultant adversarial examples suffer from poor sparsity,
making them unsuitable for interpreting the vulnerability of DNNs.

In this work, we focus on the sparse attack,
aiming to develop a new solution that yields interpretable adversarial examples,
allowing us to have a deep understanding in the vulnerability of DNNs.
To achieve our goal, we introduce a novel and \textit{theoretically solid} reparameterization technique to effectively approximate the NP-hard $l_{0}$ optimization problem,
making direct optimization of sparse perturbations
computationally tractable.
In addition, a novel loss function is proposed
to augment initial perturbations
through maximizing the adversary property 
and minimizing the number of perturbed pixels simultaneously.
As such, our approach, 
underpinned by theoretical performance guarantees,
can yield a fast, transferable, and powerful adversarial attack
while unveiling the mystery underlying adversarial perturbations.
Extensive experimental results demonstrate that our approach outperforms 
\sota\ sparse attacks in terms of computational complexity,
transferability, and attack strength.
Meanwhile, we theoretically and empirically validate 
that our approach yields much sparser adversarial examples,
suitable for interpreting the vulnerability of DNNs.
Through analyzing the minimal perturbed adversarial examples,
we discover two categories of adversarial perturbations to help understand how adversarial perturbations 
mislead the classifiers, resulted directly from
``obscuring noise'' and ``leading noise'',
where the former obscures the classifiers from identifying true classes,
while the latter misleads the classifiers into targeted predictions.

\section{Related Work}
\label{sec:rw}

The state-of-the-art solutions for adversarial attacks %,
can be grouped into two categories, 
\ie\ dense attack and sparse attack.
We shall discuss how our work relates to, and differs from, prior solutions.

\smallskip\noindent
{\bf Dense attacks} optimize the magnitude of adversarial perturbations 
under the $l_{2}$ or $l_{\infty}$ constriant.
Popularized by Fast Gradient Sign Method (FGSM)~\cite{goodfellow2015explaining}, 
FGSM-based methods are the most prominent dense attacks,
where adversarial examples~(under the $l_{\infty}$ constraint)
are effectively produced by adding the gradients of the classification loss to natural images.
Subsequent work includes I-FGSM~\cite{kurakin2017adversarial} which applies FGSM in multiple rounds,
R-FGSM~\cite{tramer2018ensemble} which augments FGSM with random initialization,
PGD~\cite{madry2018towards} which extends I-FGSM with multiple random restarts,
MI-FGSM~\cite{dong2018boosting} which boosts I-FGSM with momentum,
and DI-FGSM~\cite{xie2019DI-FGSM} and TI-FGSM~\cite{dong2019TI-FGSM}
which improve transferability respectively with random resizing and translation operations.
Other dense attacks include~\cite{szegedy2014intriguing,mohsen2016deepfool,carlini2017towards,lin20:iclr20:more_data,chen20hopskip,min22:uai:curious,fudong:cikm24:mat},
which perform effective dense attacks by
minimizing the $l_2$~(or $l_{\infty}$) distance between natural images and adversarial examples.
However, this category of solutions requires modification of almost all pixels, 
infeasible to be used for interpreting the vulnerability of DNNs.
Our solution, by contrast, perturbs only a few pixels,
able to provide additional insights about adversarial vulnerability.

\smallskip\noindent
{\bf Sparse attacks} minimize the magnitude of perturbations 
under the $l_{0}$ (or sometimes $l_{1}$) constraint.
Previous sparse attacks include 
C\&W $L_{0}$ attack~\cite{carlini2017towards} which iteratively fixes less important pixels,
OnePixel~\cite{su2019one} which applies an evolutionary algorithm, 
SparseFool~\cite{modas2019sparse} which converts 
the $l_{0}$ optimization problem to the $l_{1}$ constraint one, 
GreedyFool~\cite{dong2020greedyfool} which uses a two-stage greedy strategy, 
Homotopy attack~\cite{zhu2021homotopy} which utilizes a homotopy algorithm
to jointly optimize the sparsity and the perturbation bound,
among many others~\cite{papernot2016JSMA,xu19structured,croce19sparse,fan20factor,maura:nips21:fmn,croce:aaai22:sparse_rs,williams:cvpr23:sa_moo,ming:cvpr24:egs_tssa,vo:iclr24:brusle_attack}.
Unfortunately, prior sparse attacks suffer from 
considerable computational overhead, poor transferability, and weak attack intensity.
In contrast, we propose a theoretical sound reparameterization technique
to approximate the NP-hard $l_{0}$ optimization problem
and a novel loss function
to augment initial perturbations.
As such, our approach advances existing sparse attacks in terms of 
computational efficiency, transferability, and attack strength. 
In addition, we theoretically and empirically validate 
that our approach yields much sparser adversarial examples,
empowering us to interpret the vulnerability of DNNs.

\section{Preliminary} \label{sec:preliminary}

Given a well-trained classifier $f_{\bm{\theta}}$, %with parameters $ \bm{\theta} $,
the cross-entropy loss function $J$, and
a natural image $\bm{x}$ with the ground-truth label $y_\textrm{true}$,
the adversary aims to mislead the classifier $f_{\bm{\theta}}$ into an incorrect prediction 
via adding certain perturbation $\bm{\delta}$ 
(under the constraint $\epsilon$)
into the natural image, mathematically expressed as follows: 
\begin{equation} \label{eq:objective}
    \begin{gathered}
        f_{\bm{\theta}}(\bm{x} + \bm{\delta}) = y_\textrm{adv} 
        \quad \text{s.t.} ~~\norm{\bm{\delta}}_{p} \leq \epsilon 
        ~~\text{and}~~ \bm{x} + \bm{\delta} \in [0, 1]^\textrm{d},
    \end{gathered}
\end{equation}
where $y_\textrm{adv}$ represents the adversarial label and 
is different from $y_\textrm{true}$,
$\norm{\cdot}_{p}$ denotes the $l_{p}$ norm
(with $p= 0, 1, 2,$ or $\infty$),
and manipulating a natural image should yield one valid image.
Note here the pixel values are normalized over $[0, 1]$ to simplify the calculation.

\par\smallskip\noindent
{\bf FGSM}~(Fast Gradient Sign Method)~\cite{goodfellow2015explaining}
aims to mislead classifiers to predict incorrectly through adding gradients to natural images.
The sign function is %typically 
leveraged to ensure the perturbation $ \bm{\delta}$ 
under the $l_{\infty}$ constraint of $\epsilon$, \ie\ 
\begin{equation} \label{eq:fgsm}
\begin{gathered}
    \bm{\delta} = \epsilon \cdot \textrm{sign} (\nabla_{\bm{x}}~J(\bm{x}, ~y_\textrm{true}))  
    \quad \text{s.t.} ~~f_{\bm{\theta}}(\bm{x} + \bm{\delta}) = y_\textrm{adv}
    ~~\text{and}~~ \bm{x} + \bm{\delta} \in [0, 1]^d.
\end{gathered}
\end{equation}
FGSM is deemed as the fastest attack algorithm~\cite{dong2018boosting}.
Subsequent solutions \cite{kurakin2017adversarial,madry2018towards,xie2019DI-FGSM,dong2019TI-FGSM} have been proposed, yielding fast and highly transferable dense attacks.

\par\smallskip\noindent
{\bf I-FGSM} (Iterative Fast Gradient Sign Method)
\cite{kurakin2017adversarial} augments FGSM to have a stronger white-box attack by 
applying a small step length $\alpha$ to FGSM iteratively: 
\begin{equation} \label{eq:i-fgsm-non-targeted}
    \bm{x}^{\textrm{adv}}_{0} = \bm{x}, 
    ~~ \bm{x}^{\textrm{adv}}_{N+1} = \bm{x}^{\textrm{adv}}_{N} 
    + \alpha \cdot \textrm{sign} (\nabla_{\bm{x}}~J(\bm{x}, ~y_\textrm{true})).
\end{equation}
We can simply set $\alpha=\epsilon/T$~($T$ is the number of iterations)
to satisfy the $l_{\infty}$ constraint $\epsilon$.
Note that \eqref{eq:i-fgsm-non-targeted} 
performs a non-target attack
by adding positive gradients to natural images $ \bm{x} $
for maximizing the classification loss.
To perform a targeted attack,
we can maximize the logical probability of $y_\textrm{adv}$
on natural image $\bm{x}$ (\ie\ $\log p(y_\textrm{adv} | \bm{x})$)
by iteratively moving towards the direction 
of $\textrm{sign} \{ \nabla_{ \bm{x}}~ \log p(y_\textrm{adv} | \bm{x}) \}$:
\begin{equation} \label{eq:i-fgsm-targeted}
    \bm{x}^{\textrm{adv}}_{0} = \bm{x}, 
    ~~ \bm{x}^{\textrm{adv}}_{N+1} = \bm{x}^{\textrm{adv}}_{N} 
    - \alpha \cdot \textrm{sign} (\nabla_{ \bm{x} }~J( \bm{x}, ~y_\textrm{adv})).
\end{equation}
The intuition behind \eqref{eq:i-fgsm-targeted} is that 
adding negative gradients to the natural image $ \bm{x} $
can make the prediction of classifier $f_{\bm{\theta}}$ 
iteratively move towards the adversarial class $y_\textrm{adv}$.
% \end{comment}
%
Notably, both \eqref{eq:i-fgsm-non-targeted} and \eqref{eq:i-fgsm-targeted} must also satisfy the constraint $\bm{x}^{\textrm{adv}}_{N+1} \in [0,1]^d$ to ensure that the resulting adversarial examples remain valid input images.

\section{Our Approach} 
\label{sec:our_approach}

\subsection{Problem Statement}
\label{sec:ps}

Sparse attacks optimize the magnitude of perturbations 
under the $l_{0}$ constraint, aiming to achieve successful attacks with a small number of perturbed pixels, as formulated below:
\begin{equation} \label{eq:l0-op}
    \begin{gathered}
    \text{minimize}~\norm{\bm{\delta}}_{0}  
    \quad \text{s.t.}~~ f_{\bm{\theta}}(\bm{x} + \bm{\delta}) = y_{\textrm{adv}}
    ~\text{and}~ \bm{x} + \bm{\delta} \in [0, 1]^\textrm{d}. 
    \end{gathered}  
\end{equation}
Unfortunately, \eqref{eq:l0-op} is an NP-hard problem. When solving it, 
prior sparse attacks often got trapped in local optima~\cite{croce19sparse,zhu2021homotopy}, 
causing the resulting attacks to suffer from poor sparsity, 
unable to be used for interpreting the vulnerability of DNNs.
To address this limitation,
we develop a sparse attack that yields highly sparse adversarial examples,
suitable for understanding the vulnerability of DNNs.
% Meanwhile, it enjoys fast convergence,
% high transferability, and powerful attack strength.

% figure
\begin{figure}[!t]
    \captionsetup[subfigure]{justification=centering}
    \centering
    \begin{subfigure}[t]{0.25\textwidth}
        \centering
         \includegraphics[width=\textwidth]{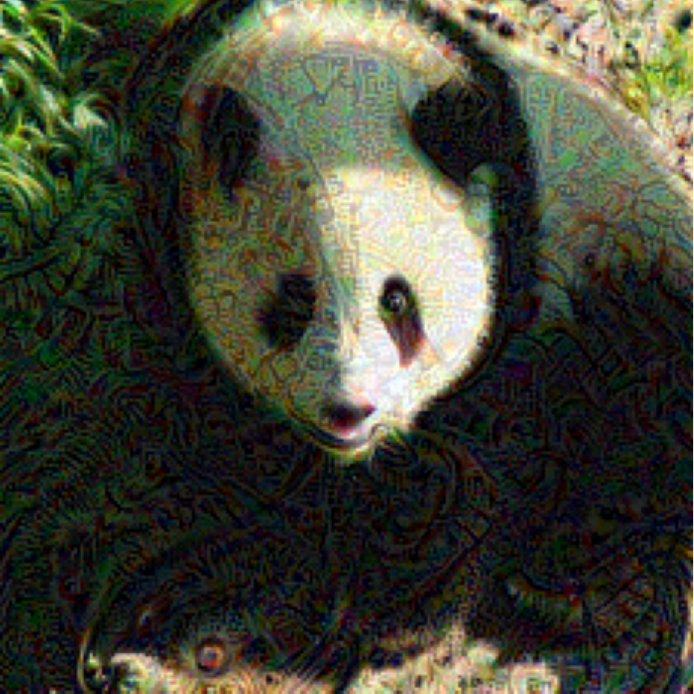}
        \caption{I-FGSM \\ ``mink'' \\  ~$98.6\% $ confidence}
        \label{fig:method-i-fgsm}
    \end{subfigure}
    \centering
     \begin{subfigure}[t]{0.25\textwidth}
        \centering
         \includegraphics[width=\textwidth]{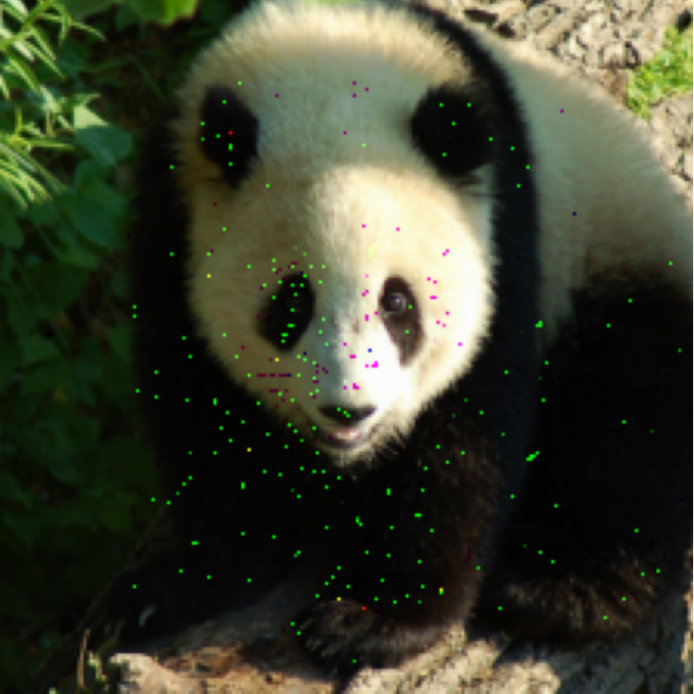}
        \caption{BruSLeAttack \\ ``panda'' \\ $13.5\% $ confidence }
        \label{fig:method-sparsefool}
    \end{subfigure}
    \centering
    \begin{subfigure}[t]{0.25\textwidth}
        \centering
         \includegraphics[width=\textwidth]{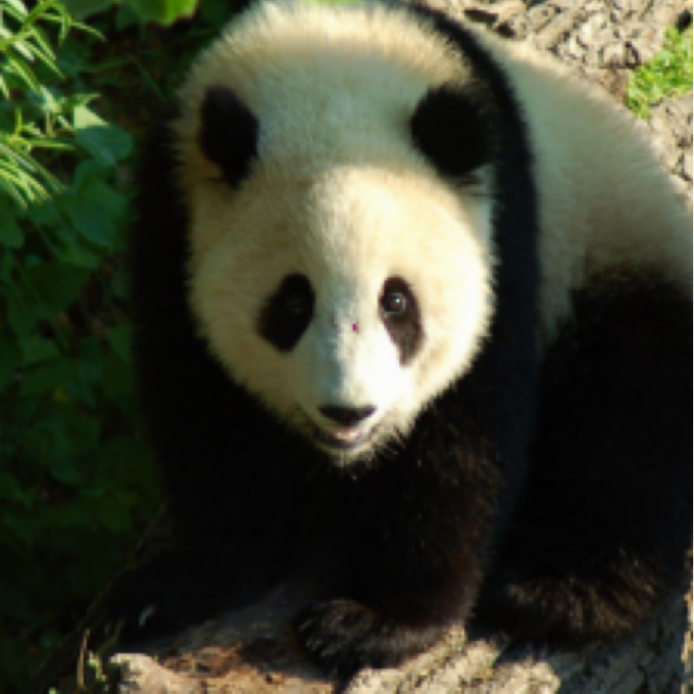}
        \caption{Ours \\ ``mink'' \\ $97.9\% $ confidence}
        \label{fig:method-ours}
    \end{subfigure}   
    
    \centering
    \begin{subfigure}[t]{0.25\textwidth}
        \centering
         \includegraphics[width=\textwidth]{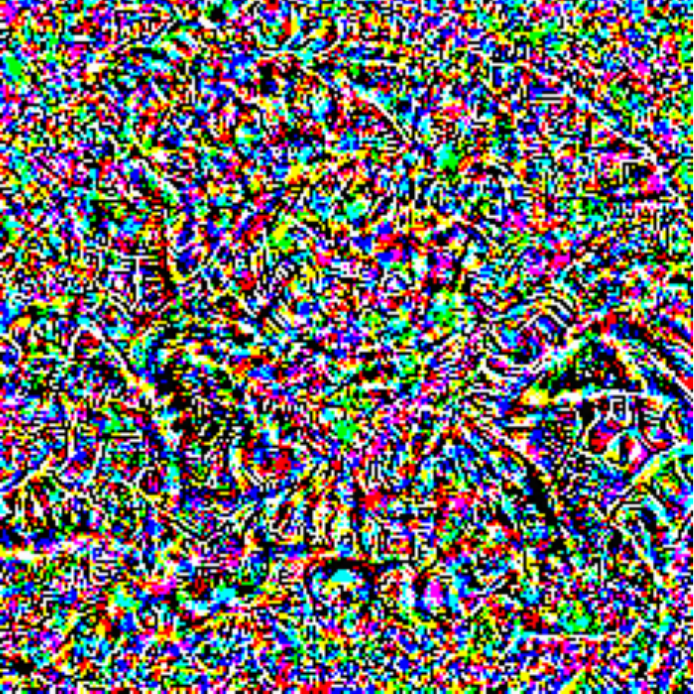}
        \caption{perturbation \\ 
        $ \norm{ \bm{\delta} }_{0} = 150523$ \\
        $(99.99 \%)$ \\
        $0.52$ seconds}
        \label{fig:method-i-fgsm-noise}
    \end{subfigure}
     \begin{subfigure}[t]{0.25\textwidth}
        \centering
          \includegraphics[width=\textwidth]{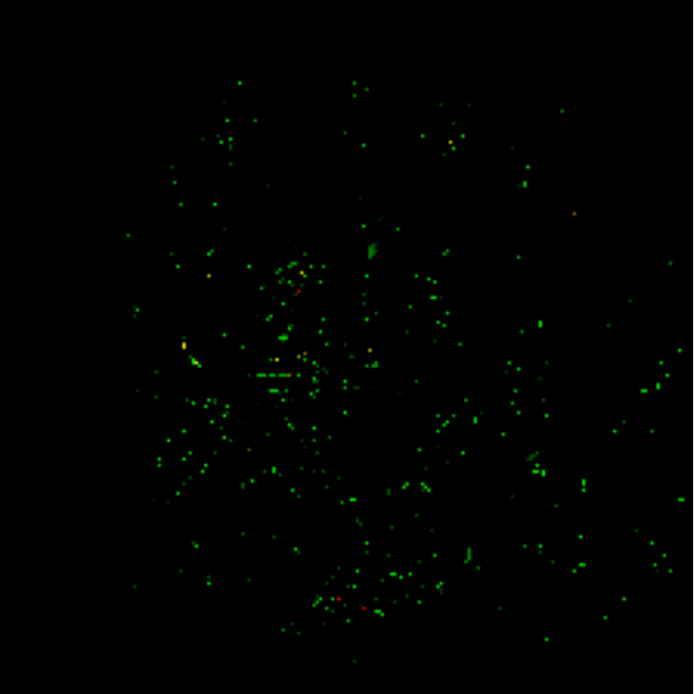} 
        \caption{
            perturbation \\ 
            $ \norm{ \bm{\delta} }_{0} = 425$ \\
            $(0.282 \%)$ \\
            $36.9$ seconds}
        \label{fig:method-sparsefool-noise}
    \end{subfigure}
    \begin{subfigure}[t]{0.25\textwidth}
        \centering
         \includegraphics[width=\textwidth]{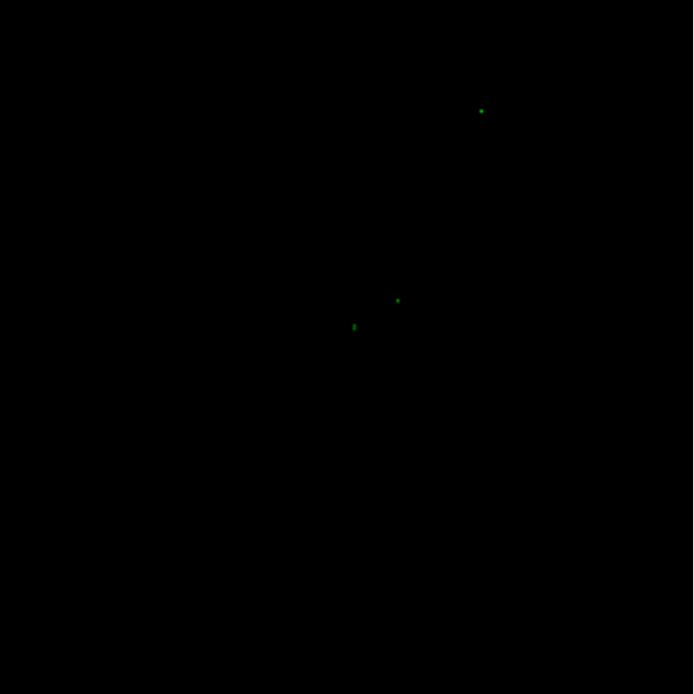}
        \caption{
            perturbation \\ 
            $ \norm{ \bm{\delta} }_{0} = 4$ \\
            $(0.003 \%)$ \\
            $3.86$ seconds}
        \label{fig:method-ours-noise}
    \end{subfigure}
    %
    % \vspace{-0.5em}
    \caption{
        Illustration of adversarial examples (AEs) computed by different attack algorithms.
        (a), (b), and (c) show AEs computed by I-FGSM, 
        BruSLeAttack, and our approach, respectively,  dependent on the classifier VGG-16.
        (d), (e), and (f) exhibit the adversarial perturbation with respect to (a), (b), and (c).
        The classification results, \ie\ ``mink'', ``panda'', ``mink'' (below (a), (b), and (c), respectively) 
        and classification confidence levels, are reported by VGG-19.
        The number of  ($\norm{ \bm{\delta}}_{0}$)
        and the percentage of perturbed pixels, as well as the computation time, 
        are all listed below (d), (e), and (f).
%        %
    }
    % \vspace{-1.0 em}
    \label{fig:method}
\end{figure}

\subsection{Challenges}
\label{sec:challenge}
Several challenges are to be addressed, as elaborated below. 

\par\smallskip\noindent
{\bf Box Constraint.}
``Box constraint'' is fundamental to adversarial attacks.
 It ensures that manipulating natural images should yield valid images 
 (\ie\ $\bm{x} + \bm{\delta} \in [0, 1]^\textrm{d}$).
 Two strategies exist to meet the box constraint for sparse attacks:
 \romsm{1}) \textit{clipping invalid pixels} which exceed the valid range 
 and  \romsm{2}) \textit{changing the optimization direction} to make  resulting adversarial examples  valid.
 However,  the former results in a severe reduction in the fooling rate, 
 while the latter incurs a large computational burden.
 So far, how to effectively handle the ``box constraint'' 
 with desired properties remains challenging. 

\par\smallskip\noindent
{\bf Computational efficiency.}   
Optimizing the magnitude of perturbations under the $l_{0}$ constraint is NP-hard.
Prior sparse attacks attempt different approximation algorithms
to reduce the computational burden,
\eg\ BruSLeAttack~\cite{vo:iclr24:brusle_attack}. 
%converts \eqref{eq:l0-op} to an $l_{1}$ constraint problem.
%
Yet, considerable computational overhead still incurs
(see Figure~\ref{fig:method-sparsefool-noise}, where $36.9$ seconds are taken). 
Hence, how to efficiently minimize the number of perturbed pixels remains open.

\par\smallskip\noindent
{\bf Transferability and Attack Strength.}
Prior sparse attacks often suffer from poor transferability and weak attack intensity.
For example, as shown in Figure~\ref{fig:method-sparsefool},
BruSLeAttack fails to perform a black-box attack,
\ie\ the adversarial example generated by VGG-16 fails to fool VGG-19.
It is challenging to develop a transferable and powerful sparse attack
with only a small amount of perturbed pixels.

\subsection{Our Idea}
\label{sec:idea}

To overcome the aforementioned challenges, 
we aim to develop a novel sparse attack
by optimizing the magnitudes of initial perturbations 
generated by I-FGSM under the $l_{0}$ constraint.
In particular, a novel loss function is introduced
to augment initial perturbations through maximizing the adversary property
and minimizing the number of perturbed pixels simultaneously.
Two observations motivate this idea.
\textit{First}, a prior study~\cite{liu2017delving} reported
that different models tend to learn similar decision boundaries.
Hence, starting from initial perturbations generated by dense attacks
accelerates convergence and 
makes our approach more likely to reach the global optima~\cite{suya2020hybrid}.
\textit{Second}, I-FGSM results in overly perturbed adversarial examples, 
and thereby reducing the number of perturbed pixels 
may not negatively affect its adversary property.
For example, as depicted in Figure~\ref{fig:method-i-fgsm} (I-FGSM) 
and Figure~\ref{fig:method-ours} (Our Approach),
both I-FGSM and our approach mislead VGG-19 into the same incorrect prediction (\ie\ ``mink'').
Therefore, it is feasible to optimize the magnitude of 
perturbations produced by I-FGSM under the $l_{0}$ constraint 
without sacrificing its transferability.
Furthermore, a new box constraint strategy
is designed to make our solution always yield valid adversarial examples.

\subsection{Our Proposed Approach} 
\label{sec:approach}

\par\smallskip\noindent
{\bf Objective.}
To achieve our goal, we propose a novel sparse attack
%, named as \textit{Weighted Iterative Fast Gradient Sign Method} (WI-FGSM),
by optimizing the magnitude of initial perturbations under the $l_{0}$ constraint.
Our problem can be formulated as follows:
\begin{equation} \label{eq:objective}
\begin{gathered}
    \text{minimize} ~~\norm{ \bm{w} }_{0} 
    \quad \text{s.t.} ~~ f_{\bm{\theta}}(\bm{x} + \bm{w} \odot \bm{\delta}) = y_{\textrm{adv}}
    ~\text{and}~ \bm{x} + \bm{w} \odot \bm{\delta} \in [0, 1]^\textrm{d},
\end{gathered}    
\end{equation}
where $\bm{\delta}$ indicates the initial perturbation,
$\bm{w}$ denotes the weight with the same dimension as perturbation $\bm{\delta}$,
and $\odot$ represents an element-wise product.
Motivated by the ``pre-train then tune'' paradigm~\cite{zeiler2014visualize},
instead of randomly initializing the perturbation $\bm{\delta}$,
%we pre-compute it by using I-FGSM,
we pre-compute it by using \textit{Iterative Fast Gradient Sign Method} (I-FGSM)~\cite{kurakin2017adversarial},
which in turn accelerates convergence and makes the resulting perturbation more likely to 
reach the global optima~\cite{suya2020hybrid}.
Note that solving the problem formulated by~\eqref{eq:objective} yields 
near-optimal solutions for \eqref{eq:l0-op}, 
\ie\ with the fewest perturbed pixels.
%
% The underlying intuition is deferred to Section A of supplementary materials.

To augment initial perturbations,
we follow prior work~\cite{carlini2017towards,liu2017delving} to
reformulate our problem of \eqref{eq:objective} via the Lagrangian relaxation formulation
for concurrently maximizing the classification loss 
and minimizing the number of perturbed pixels,
yielding:
\begin{equation} \label{eq:loss}
    J_{\textrm{adv}} = J(f_{\bm{\theta}}(\bm{x} + \bm{w} \odot \bm{\delta}), ~y_{\textrm{adv}}) + \lambda~\norm{\bm{w}}_{0},
\end{equation}
where $\lambda$ is a hyperparameter to 
balance the classification loss and the degree of perturbation.
As negative elements in $\bm{w}$ indicate that the corresponding perturbations in $\bm{\delta}$  
negatively affect the adversary property,
a simple but effective way to augment \eqref{eq:loss} is via applying
the ReLU function to drop negative values in $\bm{w}$, \ie
\begin{equation} \label{eq:loss-with-relu}
    J_{\textrm{adv}} = J(f_{\bm{\theta}}(\bm{x} + \pi( \bm{w}) \odot \bm{\delta}), ~y_{\textrm{adv}}) + \lambda~\norm{ \pi (\bm{w} ) }_{0},
\end{equation}
where $\pi ( \cdot )$ represents the ReLU function.

\par\smallskip\noindent
{\bf Reparameterization Technique.}
However, as indicated by~\cite{louizos:iclr18:l0-regularization},
the $l_{0}$ norm of $\pi (\bm{w})$ is non-differentiable,
so directly optimizing \eqref{eq:loss-with-relu} is  computationally intractable.
To address this intractability, let $H (\cdot)$ be the Heaviside step function and
consider a simple re-parametrization technique:
\begin{equation} \label{eq:re-para}
    \begin{aligned}
        % \bm{w}^{\prime} = H \left( \pi (\bm{w}) \right), 
         \norm{ \pi (\bm{w})}_{0} = \sum_{j=1}^{d} H(\pi (w_{j}) ),
         ~\text{with}~  H(x) = \begin{cases}
            1,  & x > 0 \\
            0,  & x \leq 0 \\
        \end{cases}.
    \end{aligned}
\end{equation}
As such, we can reformulate \eqref{eq:loss-with-relu} as follows,
\begin{equation} \label{eq:loss-re-para}
    \begin{aligned}
        & J_{\textrm{adv}} = J(f_{\bm{\theta}}(\bm{x} +  \pi (\bm{w})  \odot \bm{\delta}), ~y_{\textrm{adv}}) 
        + \lambda \sum_{j=1}^{d} H \left( \pi (w_{j}) \right).
    \end{aligned}
\end{equation}
Obviously, penalizing the second term of \eqref{eq:loss-re-para} 
is equivalent to penalize the $l_{0}$ norm of $\pi ( \bm{w} )$.
However, it is impractical to directly optimize \eqref{eq:loss-re-para}
because the distributional derivative of the Heaviside step function,
\ie\ the Dirac delta function,
equals to zero almost everywhere~\cite{wiki:step_function}.
The Dirac delta function, by definition, can be simply regarded as a function 
that is zero everywhere except at the origin, where it is infinite, \ie\
\begin{equation} \label{eq:delta_func}
    \begin{aligned}
         p(x) \simeq  \begin{cases}
            \infty,  & x = 0 \\
            0,  & x \neq 0 \\
        \end{cases}.
    \end{aligned}
\end{equation}
To make the Heaviside step function differentiable,
we devise a novel reparametrization technique
to approximate the Dirac delta function,
resorting to the zero-centered normal distribution
presented as follows:
\begin{equation} \label{eq:normal_approxiate}
    \begin{aligned}
         q_{a} (x) = \frac{1}{|a|\sqrt{\pi}} \exp^{-(x/a)^{2}}.
    \end{aligned}
\end{equation}
Here, the variance of $q_{a} (x)$ is determined by the hyperparameter $a$.
When the hyperparameter $a$ approaches zero, 
the function $q_{a}(x)$ converges to the Dirac delta function $p(x)$,
as stated next.

\begin{theorem} \label{thm:convergence}
    (Convergence) 
    Let $p(x)$ denote the Dirac delta function. 
    Consider $q_{a}(x)$ defined as
    $q_{a} (x) = \frac{1}{|a|\sqrt{\pi}} \exp^{-(x/a)^{2}}$, 
    which represents a zero-centered normal distribution with variance dependent on the hyperparameter $a$. Then, in the distributional sense, we have:
    \begin{equation} \label{eq:limit}
       \lim_{a\rightarrow 0} q_{a} (x) = p(x).
    \end{equation}
\end{theorem}

\begin{proof}
    First, consider the case when $x=0$, 
    we always have:
    \begin{equation} 
    \label{eq:thm_case1}
        \lim_{a\rightarrow 0} q_{a} (x) 
        = \lim_{a\rightarrow 0} \frac{1}{|a|\sqrt{\pi}} \exp^{-(0/a)^{2}}
        = \lim_{a\rightarrow 0} \frac{1}{|a|\sqrt{\pi}} = \infty.
    \end{equation}
    Second, when $x \neq 0$,
    we need to consider both the positive and negative directions of $a$, \ie\
    $\lim_{a\rightarrow 0^{+}}$ 
    and $\lim_{a\rightarrow 0^{-}}$.
    Starting with the positive direction,
    let $t = \frac{1}{a}$, 
    we have 
    \begin{equation} 
    \label{eq:thm_case2}
        \begin{gathered}
            \lim_{a\rightarrow 0^{+}} q_{a} (x) 
            = \lim_{a\rightarrow 0} \frac{1}{a\sqrt{\pi}} \exp^{-(x/a)^{2}}
            = \lim_{t \rightarrow \infty} \sqrt{\pi} t \cdot \exp^{-x^{2}t^{2}} 
            = \lim_{t \rightarrow \infty} \frac{\sqrt{\pi} t}{\exp^{x^{2} t^{2}}} \\
            = \lim_{t \rightarrow \infty} \frac{\sqrt{\pi}}{2x^{2}t \cdot \exp^{x^{2} t^{2}}} ~\text{(L'Hopital's rule)}
            = \frac{\sqrt{\pi}}{\infty} = 0. 
        \end{gathered}
    \end{equation}
    Similarly, for the negative direction of $a$,
    we have:
    \begin{equation} 
    \label{eq:thm_case2}
        \begin{gathered}
            \lim_{a\rightarrow 0^{-}} q_{a} (x)
            = \lim_{a\rightarrow 0} -\frac{1}{a\sqrt{\pi}} \exp^{-(x/a)^{2}}
            = \lim_{t \rightarrow \infty} -\sqrt{\pi} t \cdot \exp^{-x^{2}t^{2}} \\
            = \lim_{t \rightarrow \infty} - \frac{\sqrt{\pi} t}{\exp^{x^{2} t^{2}}} 
            = \lim_{t \rightarrow \infty} - \frac{\sqrt{\pi}}{2x^{2}t \cdot \exp^{x^{2} t^{2}}}
            = - \frac{\sqrt{\pi}}{\infty} = 0. 
       \end{gathered}
    \end{equation}
    
    Based on the above discussion, we have: 
    \begin{equation} \label{eq:limit_proof}
       \lim_{a\rightarrow 0} q_{a} (x) = \begin{cases}
            \infty,  & x = 0 \\
            0,  & x \neq 0 \\
        \end{cases}
        = p(x).
    \end{equation}
    
\end{proof}

% derivative
\begin{figure} [!t] %[htbp]
    \centering
    \includegraphics[width=.55\textwidth]{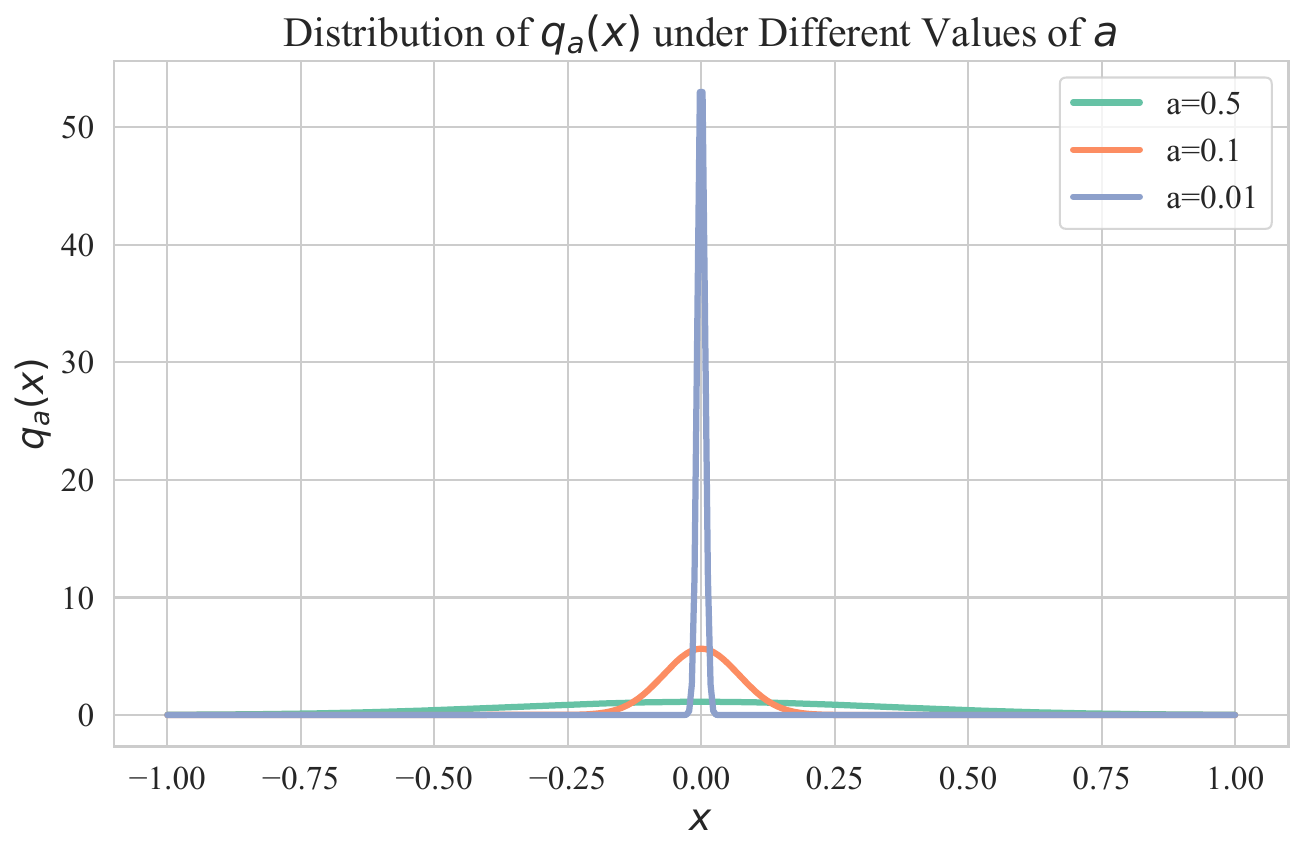}
      % \vspace{-0.5 em}
    \caption{
        Distribution of $q_{a}(x)$ under different values of $a$. 
        As $a$ approaches to $0$, 
        $q_{a} (x)$ increasingly resembles the Dirac delta function.
    }
    \label{fig:method-normal-dist}
    % \vspace{-2.0 em}
\end{figure}

% Due to the page limit, the proof of Theorem~\ref{thm:convergence} is deferred to Appendix A.1.
%
Figure~\ref{fig:method-normal-dist} presents the distribution of $q_{a}(x)$ for various values of $a$. 
It is clear that as $a$ approaches zero, 
the shape of $q_{a}(x)$ more closely resembles the Dirac delta function, a crucial aspect in estimating the derivative of the Heaviside step function.
This estimation is significant as it allows for the differentiability of \eqref{eq:loss-re-para}, demonstrated by:

\begin{equation} \label{eq:derivetive}
    \begin{aligned}
        \frac{dH}{dx} \approx \frac{1}{|a|\sqrt{\pi}} \exp^{-   (x/a)^{2}}.
    \end{aligned}
\end{equation}
Here, the hyperparameter $a$ modulates the balance between
optimization smoothness and approximation accuracy.

\par\smallskip\noindent
{\bf Performance Guarantees.}
%In addition, a 
A simple trick to increase the sparsity for convergence acceleration 
is to tailor the ReLU function as follows:
\begin{equation} \label{eq:convergence}
    \pi^{\prime} (\bm{w}) = \pi (\bm{w} - \frac{\tau}{\epsilon}),
\end{equation}
where $\epsilon$ is the $l_{\infty}$ constraint for I-FGSM
and $\tau$ is used to simplify hyperparameter tuning.
By using our tailored ReLU function $\pi^{\prime} (\cdot)$,
we can re-write the loss function given below:
%
% final loss
\begin{equation} \label{eq:final-loss}
    \begin{gathered}
         J_{\textrm{adv}} = J(f_{\bm{\theta}}(\bm{x} +  \pi (\bm{w} - \frac{\tau}{\epsilon} )  \odot \bm{\delta}), ~y_{\textrm{adv}}) 
        + ~\lambda \sum_{j=1}^{d} H \left( \pi (w_{j} - \frac{\tau}{\epsilon} ) \right). 
    \end{gathered}
\end{equation}
By employing \eqref{eq:final-loss},
our approach yields much sparser adversarial examples,
as stated below.
\begin{prop} \label{thm:sparsity}
    (Sparsity) 
    Given an initial perturbation $ \bm{\delta} \in [-\epsilon, \epsilon]^{d}$
    under some $l_{\infty}$ constraint $\epsilon~(\epsilon > 0) $,
    let $\bm{w} \in \mathbb{R} ^{d} $ be the weight matrix used in our study,
    and $\pi( \cdot)$ be the ReLU function, 
    in terms of the $l_{0}$ norm, we have:
    \begin{equation} \label{eq:sparsity}
       \norm{ \pi (\bm{w} - \frac{\tau}{\epsilon} ) \odot \bm{\delta}}_{0} 
       ~\leq~ \norm{  \pi (\bm{w}) \odot \bm{\delta} }_{0}
       ~\leq~ \norm{ \bm{w} \odot \bm{\delta} }_{0}.
    \end{equation}
\end{prop}
The proof of Proposition~\ref{thm:sparsity} is deferred to Appendix A.1 for conserving sparse.

Although \eqref{eq:final-loss} can yield more sparse adversarial examples,
it cannot guarantee valid pixel values,
\ie\ $\exists ~w_{j}  \in \bm{w}, 
x_{j} +  \pi (w_{j} - \frac{\tau}{\epsilon} ) \odot \delta_{j} \notin [0, 1]$.
To remedy this,  we devise a new strategy to satisfy the box constraint:
\begin{equation} \label{eq:bound}
    \bm{w}^{\prime} = \bm{\Omega} \odot H \left( \pi (\bm{w} - \frac{\tau}{\epsilon} ) \right),
\end{equation}
where $\bm{\Omega}$ serves to impose a tight bound on the resulting perturbation.
Following this step, we are able to 
obtain valid adversarial examples, as stated next.
\begin{prop} \label{thm:bound}
     (Box constraint)
    Let $\bm{x} \in [0, 1]^\textrm{d}$ be a natural image, 
    $\bm{\delta} \in [-\epsilon, \epsilon]^{d}$ be the initial perturbation 
    % computed by I-FGSM 
    under a $l_{\infty}$ constraint $\epsilon$ ($\epsilon > 0$), 
    and $\bm{w}^{\prime}$ be the output of \eqref{eq:bound}.
    If ~$\bm{\Omega} = \min(\frac{\bm{x}}{\epsilon}, \frac{1-\bm{x}}{\epsilon})$,
    the resulting adversarial example is valid, \ie\ 
    \begin{equation} \label{eq:thm}
        \bm{x} + \bm{w}^{\prime} \odot \bm{\delta} \in [0, 1]^{\textrm{d}}.
    \end{equation}
    %\vspace{-1.0 em}
\end{prop}

The proof of Proposition~\ref{thm:bound} is deferred to Appendix A.2.
Note that such a box constraint strategy can automatically yield valid pixels, 
making adversarial attacks escape from local optima, 
as stated in the prior study~\cite{carlini2017towards}.
Details of our algorithm flow are deferred to Appendix B.

\section{Experiments and Results}
\label{sec:exp}

\subsection{Experimental Settings} 
\label{sec:exp-setup}

\par\smallskip\noindent
{\bf Datasets.} We experimentally benchmark our approach 
on three widely used image  datasets: \romsm{1}) 
$70,000$ greyscale examples of \textbf{MNIST}~\cite{lecun2010mnist}; 
\romsm{2})  $60,000$ RGB examples of \textbf{CIFAR-10}~\cite{krizhevsky2009cifar10};
\romsm{3})  $10,000$ RGB examples randomly selected from the \textbf{ImageNet}~\cite{russakovsky2015imagenet}
validation set.

\par\smallskip\noindent
{\bf Compared Methods.} We compare our approach to ten sparse attack counterparts,
\ie\ \textbf{C\&W} \bm{$L_{0}$}~\cite{carlini2017towards}, \textbf{OnePixel}~\cite{su2019one}, \textbf{SparseFool}~\cite{modas2019sparse}, 
\textbf{GreedyFool}~\cite{dong2020greedyfool}, 
\textbf{FMN}~\cite{maura:nips21:fmn},
\textbf{Homotopy}~\cite{zhu2021homotopy},
\textbf{Sparse-RS}~\cite{croce:aaai22:sparse_rs},
\textbf{SA-MOO}~\cite{williams:cvpr23:sa_moo},
\textbf{EGS-TSSA}~\cite{ming:cvpr24:egs_tssa}
and \textbf{BruSLeAttack}~\cite{vo:iclr24:brusle_attack}.
Hyperparameters for compared methods, if not specified, 
are set as mentioned in their respective articles.
All comparative results represent the average of $5$ trials.

\par\smallskip\noindent
{\bf Parameter Settings.} 
Four pre-trained models, 
VGG-16~\cite{simonyan2014very}, VGG-19, ResNet-101~\cite{he2016deep}, and ResNet-152,
are exploited to evaluate the adversarial perturbation under ImageNet. 
The hyperparameters for I-FGSM, unless specified otherwise, are set as $\epsilon=4/255$, $\alpha=1/255$, and the number of iterations equal to $10$.
We optimize the weight in our approach by using SGD 
with a momentum of $0.9$ and a learning rate of $1e-2$.
A mini-batch of $256$ is used for MNIST and CIFAR-10, 
and of $64$ for ImageNet.
We set $\lambda=1e-2$ for MNIST and CIFAR-10, 
and $\lambda=1e-3$ for ImageNet.
We grid-search $a$ (and $\tau$) and empirically set them to $0.1$ (and $0.30$)
for all three datasets.
The number of iterations for our approach is set to $100$, $100$, and $200$ for MNIST, CIFAR-10, and ImageNet, respectively.
For the targeted attack, %if not specified,
we follow prior studies~\cite{kurakin2017adversarial,kurakin2017scale}
by setting the least-likely class as the targeted label.
All experiments were conducted on a workstation equipped with an RTX 4090 GPU.

\subsection{Evaluation on Sparsity} \label{sec:sparsity}
    We compare our approach to sparse attacks listed in Section~\ref{sec:exp-setup}
    in terms of sparsity (\ie\ $l_{0}$ norm of perturbation)
    under the non-targeted and targeted attack scenarios.
    ResNet-18 and VGG-16 are used to generate adversarial examples 
    and report the classification results for CIFAR-10 and ImageNet, respectively.
    For all methods (except for OnePixel),
    we report the averagely required amount of perturbed pixels for achieving a fooling rate of $100 \%$ in both scenarios.
    Note that OnePixel cannot achieve such a fooling rate 
    because its perturbations are limited to an extreme case. 
    Table~\ref{tab:sparsity-targeted} presents the experimental results.
    Under the non-targeted attack scenario, 
    we observe that except for OnePixel, 
    our approach achieves the minimal magnitude of perturbations,
    with the averaged number of $44$ ($1.45 \%$ pixels) and of $57$ ($0.04 \%$ pixels) for CIFAR-10 and ImageNet, respectively.    
    Although OnePixel outperforms our approach in terms of sparsity,
    it suffers from an extremely poor fooling rate,
    substantially inferior to our approach.
    Similarly, under the targeted attack scenario,
    our approach achieves the best sparsity of $69$ ($2.27 \%$ pixels) and of $136$ ($0.09 \%$ pixels)
    for CIFAR-10 and ImageNet, respectively, significantly outperforming all its counterparts.
    Notably, OnePixel and SpareFool cannot perform effective targeted attacks, so their results are unavailable.

% sparsity
% \input{C-1-table-sparsity-target.tex}

%
\begin{table} [!t]  %[htbp] %[!t] 
    % \tiny
    % \small
   \scriptsize
   \centering
   \setlength\tabcolsep{5 pt}
  \caption{Comparative results of sparsity under CIFAR-10 and ImageNet,
  with the minimum amounts of perturbed pixels required to achieve the fooling rate (FR) of $100 \%$ for non-targeted and targeted attacks reported
  }
  % \vspace{-0.5 em}
  \begin{tabular}{@{}|c|cccc|cccc|@{}}
   \toprule
   \multirow{3}{*}{Method} & \multicolumn{4}{c|}{Non-targeted Attack}                                                                                  & \multicolumn{4}{c|}{Targeted Attack}                                                                             \\ \cmidrule(l){2-9} 
                           & \multicolumn{2}{c|}{CIFAR-10}                                    & \multicolumn{2}{c|}{ImageNet}                          & \multicolumn{2}{c|}{CIFAR-10}                                                         & \multicolumn{2}{c|}{ImageNet}                                         \\ \cmidrule(l){2-9} 
                           & \multicolumn{1}{c}{FR  }                           & \multicolumn{1}{c|}{Sparsity }                          & \multicolumn{1}{c}{FR }                & Sparsity                              & \multicolumn{1}{c}{FR  }                    & \multicolumn{1}{c|}{Sparsity   }                        & \multicolumn{1}{c}{FR  }        & Sparsity     \\ \midrule
   C\&W $L_{0}$            & \multicolumn{1}{c}{$100\%$}                                      & \multicolumn{1}{c|}{$52$}                               & \multicolumn{1}{c}{$100\%$}                     & $424$                              & \multicolumn{1}{c}{$100\%$}                          & \multicolumn{1}{c|}{$102$}                             & \multicolumn{1}{c}{$100\%$}                          & $5132$                     \\ 
   OnePixel                & \multicolumn{1}{c}{$35.2\%$}                                     & \multicolumn{1}{c|}{$1$}                               & \multicolumn{1}{c}{$20.5\%$}                     & $3$                                & \multicolumn{1}{c}{-}                                & \multicolumn{1}{c|}{-}                                 & \multicolumn{1}{c}{-}                                & -                          \\ 
   SparseFool              & \multicolumn{1}{c}{$100\%$}                                      & \multicolumn{1}{c|}{$76$}                              & \multicolumn{1}{c}{$100\%$}                      & $234$                              & \multicolumn{1}{c}{-}                                & \multicolumn{1}{c|}{-}                                 & \multicolumn{1}{c}{-}                                & -                          \\ %\midrule
   FMN               & \multicolumn{1}{c}{$100\%$}                                      & \multicolumn{1}{c|}{$106$}                             & \multicolumn{1}{c}{$100\%$}                      & $632$                              & \multicolumn{1}{c}{$100\%$}                          & \multicolumn{1}{c|}{$183$}                             & \multicolumn{1}{c}{$100\%$}                          & $1087$                     \\ %\midrule
   Sparse-RS              & \multicolumn{1}{c}{$100\%$}                                      & \multicolumn{1}{c|}{$167$}                             & \multicolumn{1}{c}{$100\%$}                      & $758$                              & \multicolumn{1}{c}{$100\%$}                          & \multicolumn{1}{c|}{$244$}                             & \multicolumn{1}{c}{$100\%$}                          & $1293$                     \\ %\midrule
   SA-MOO                & \multicolumn{1}{c}{$100\%$}                                      & \multicolumn{1}{c|}{$173$}                             & \multicolumn{1}{c}{$100\%$}                      & $1392$                             & \multicolumn{1}{c}{$100\%$}                          & \multicolumn{1}{c|}{$305$}                             & \multicolumn{1}{c}{$100\%$}                          & $2896$                     \\ 
    EGS-TSSA                & \multicolumn{1}{c}{$100\%$}                                      & \multicolumn{1}{c|}{$105$}                             & \multicolumn{1}{c}{$100\%$}                      & $549$                             & \multicolumn{1}{c}{$100\%$}                          & \multicolumn{1}{c|}{$218$}                             & \multicolumn{1}{c}{$100\%$}                          & $1054$                     \\ 
    BruSLeAttack                & \multicolumn{1}{c}{$100\%$}                                      & \multicolumn{1}{c|}{$93$}                             & \multicolumn{1}{c}{$100\%$}                      & $164$                             & \multicolumn{1}{c}{$100\%$}                          & \multicolumn{1}{c|}{$148$}                             & \multicolumn{1}{c}{$100\%$}                          & $678$                     \\ \midrule
   Ours                    & \multicolumn{1}{c}{$100\%$}                                      & \multicolumn{1}{c|}{$44$}                              & \multicolumn{1}{c}{$100\%$}                      & $57$                               & \multicolumn{1}{c}{$100\%$}                          & \multicolumn{1}{c|}{$69$}                              & \multicolumn{1}{c}{$100\%$}                          & $136$                      \\ \bottomrule
   \end{tabular}
  \label{tab:sparsity-targeted}
  % \vspace{-1.0 em}
\end{table}

% sparsity - one column
\begin{figure} [!t] % [htbp] 
    % \vspace{0.5 em}
        \captionsetup[subfigure]{justification=centering}
             %fmn
            \centering
            \begin{subfigure}[t]{0.3\textwidth}
                \centering
                \includegraphics[width=\textwidth]{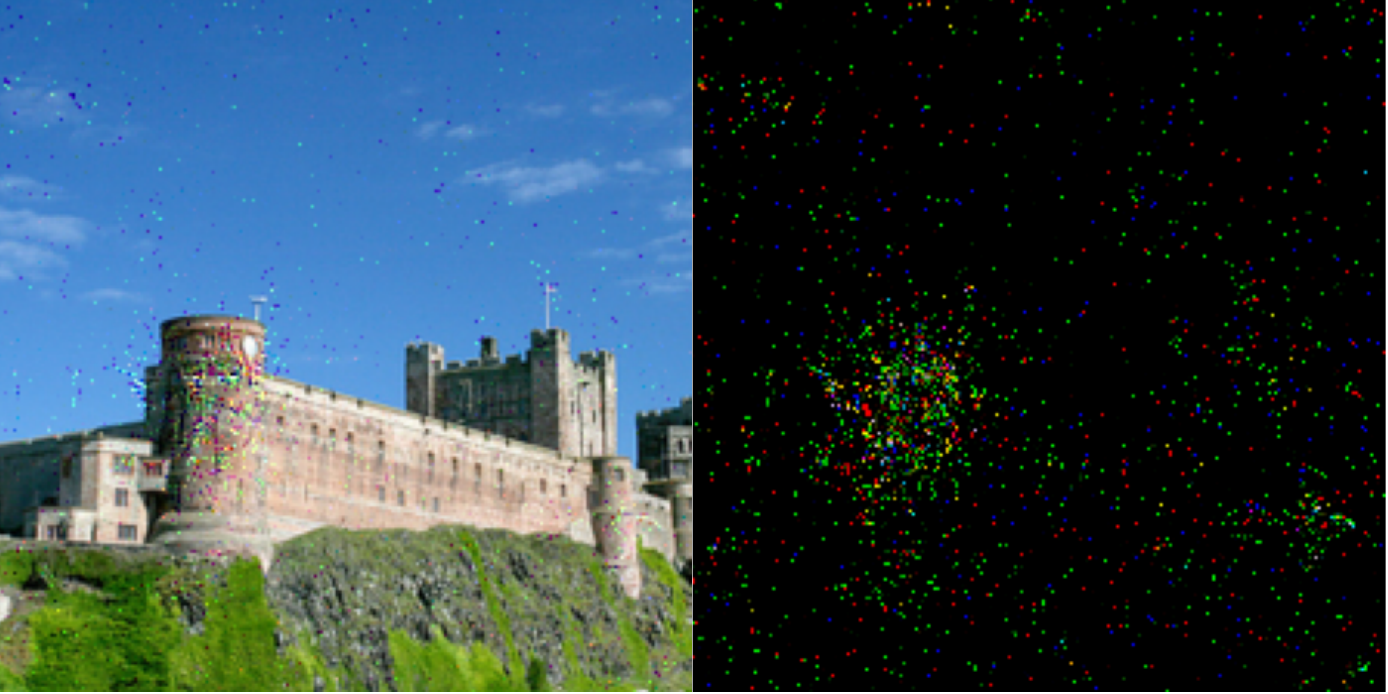}
                \caption{C\&W $L_{0}$ \\ 
                    ``wombat",~ \\
                    $ \norm{ \bm{\delta}}_{0} = 3555$  }
                \label{fig:reb-cw-castle}
            \end{subfigure}
            \centering
            \begin{subfigure}[t]{0.3\textwidth}
                \centering
                \includegraphics[width=\textwidth]{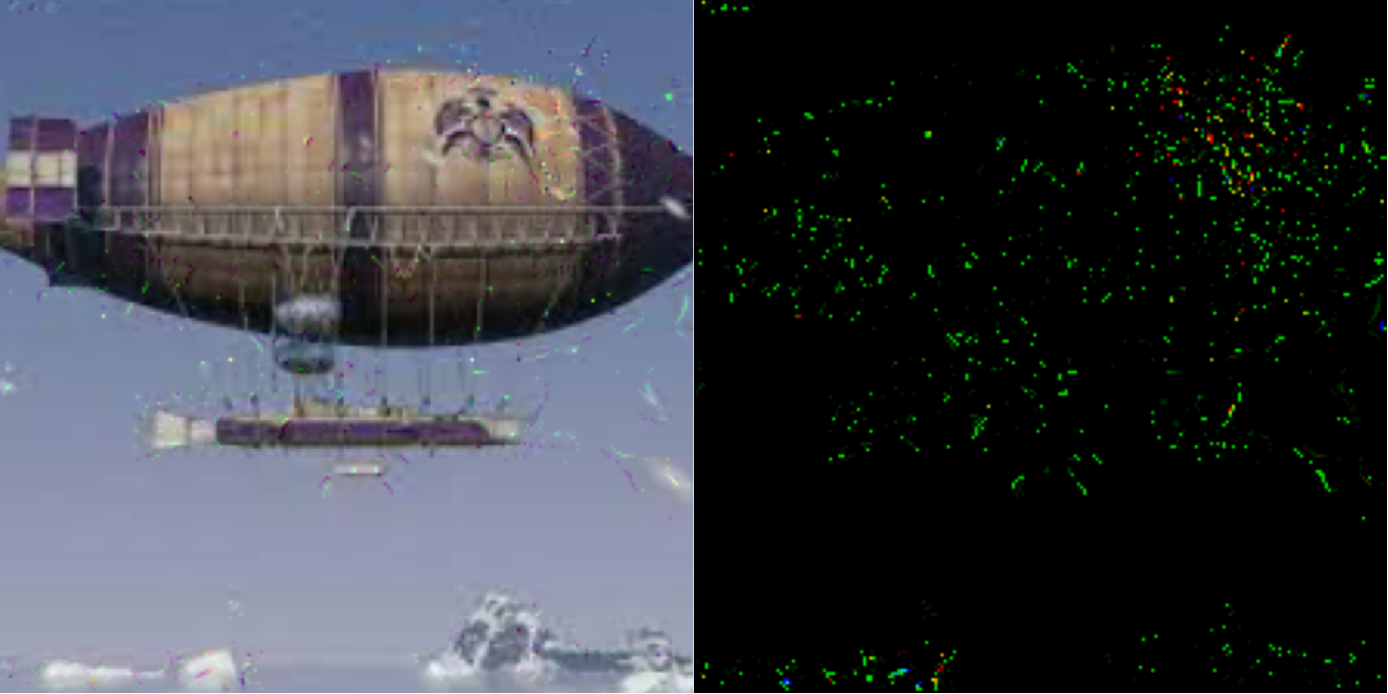}
                \caption{GreedyFool \\ 
                    ``garter snake",~ \\
                    $ \norm{ \bm{\delta}}_{0} = 1070$  }
                \label{fig:reb-fmn-airship}
            \end{subfigure}
            \centering
            \begin{subfigure}[t]{0.3\textwidth}
                \centering
                \includegraphics[width=\textwidth]{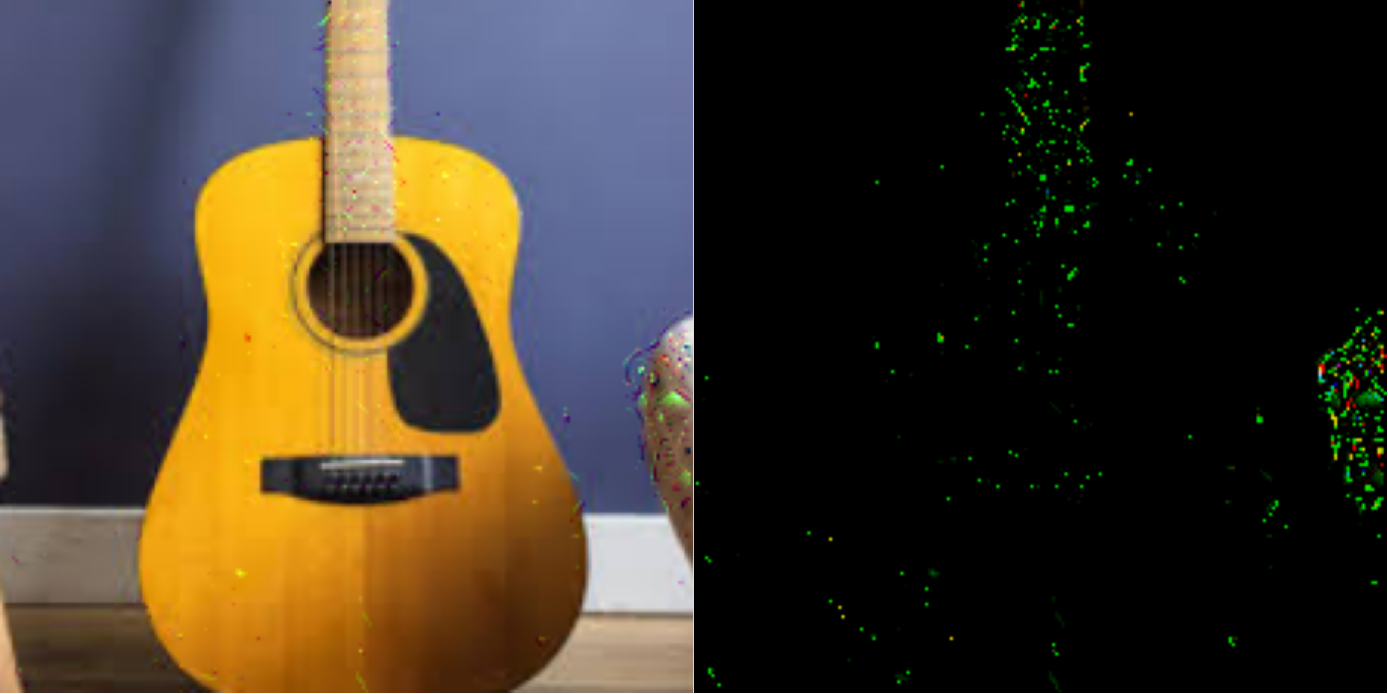}
                \caption{Homotopy \\ 
                    ``tailed frog",~ \\
                    $ \norm{ \bm{\delta}}_{0} = 1731$  }
                \label{fig:reb-fmn-guitar}
            \end{subfigure}

             \centering
             \begin{subfigure}[t]{0.3\textwidth}
                 \centering
                 \includegraphics[width=\textwidth]{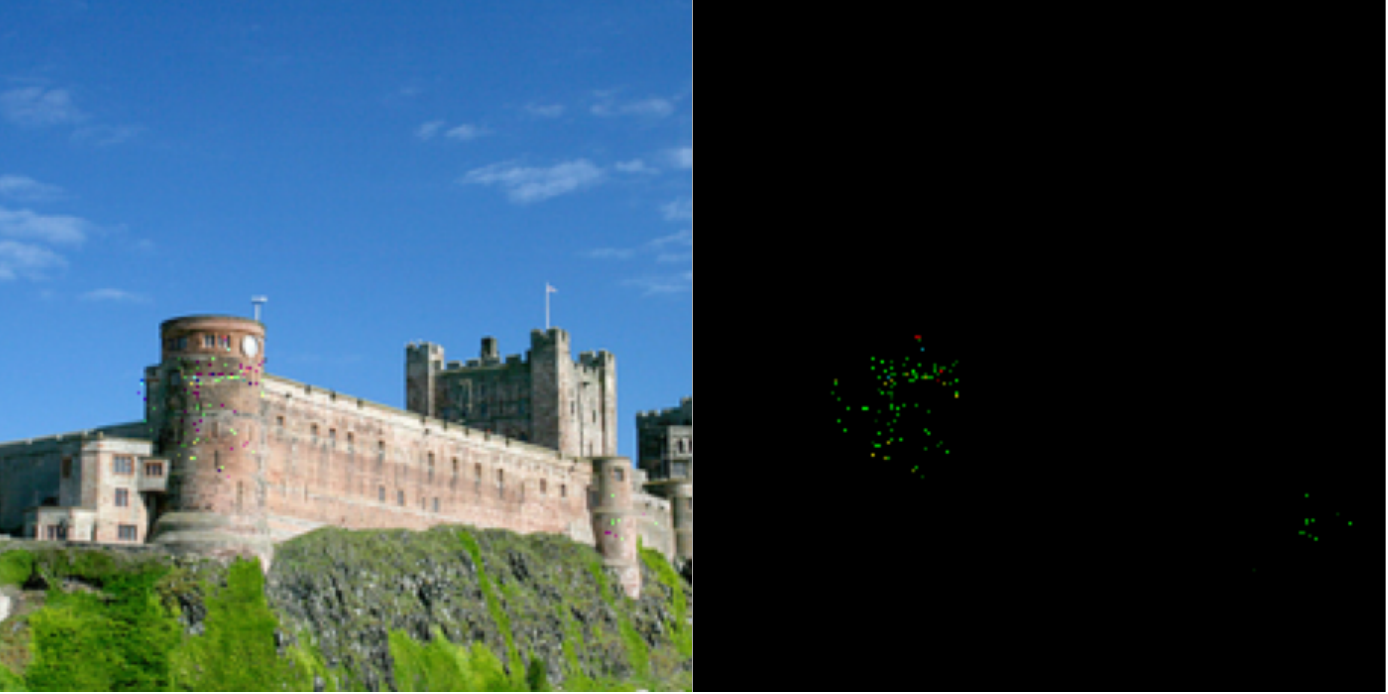}
                 \caption{Ours \\ 
                 ``wombat",~ \\
                     $ \norm{ \bm{\delta}}_{0} = 123$  }
                 \label{fig:reb-fmn-castle}
             \end{subfigure}
             \centering
             \begin{subfigure}[t]{0.3\textwidth}
                 \centering
                 \includegraphics[width=\textwidth]{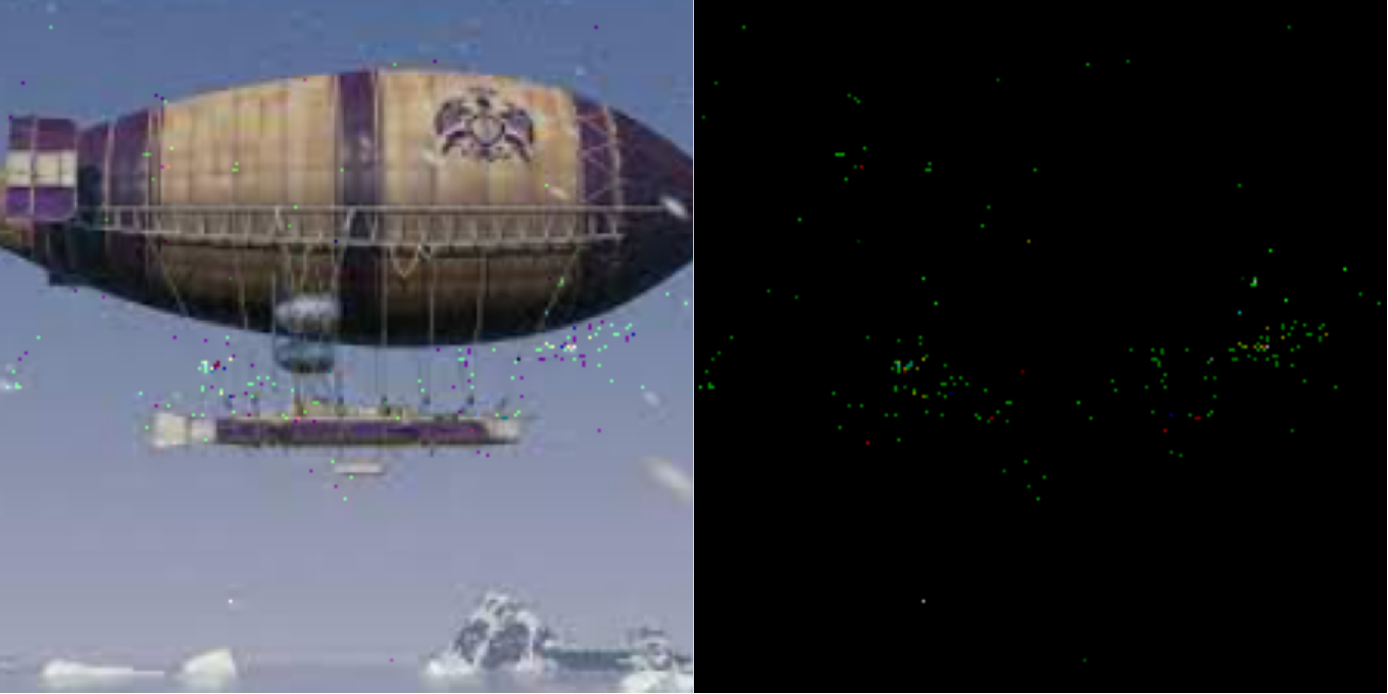}
                 \caption{Ours \\ 
                     ``garter snake",~ \\
                     $ \norm{ \bm{\delta}}_{0} = 107$  }
                 \label{fig:reb-fmn-airship}
             \end{subfigure}
            \centering
            \begin{subfigure}[t]{0.3\textwidth}
                \centering
                \includegraphics[width=\textwidth]{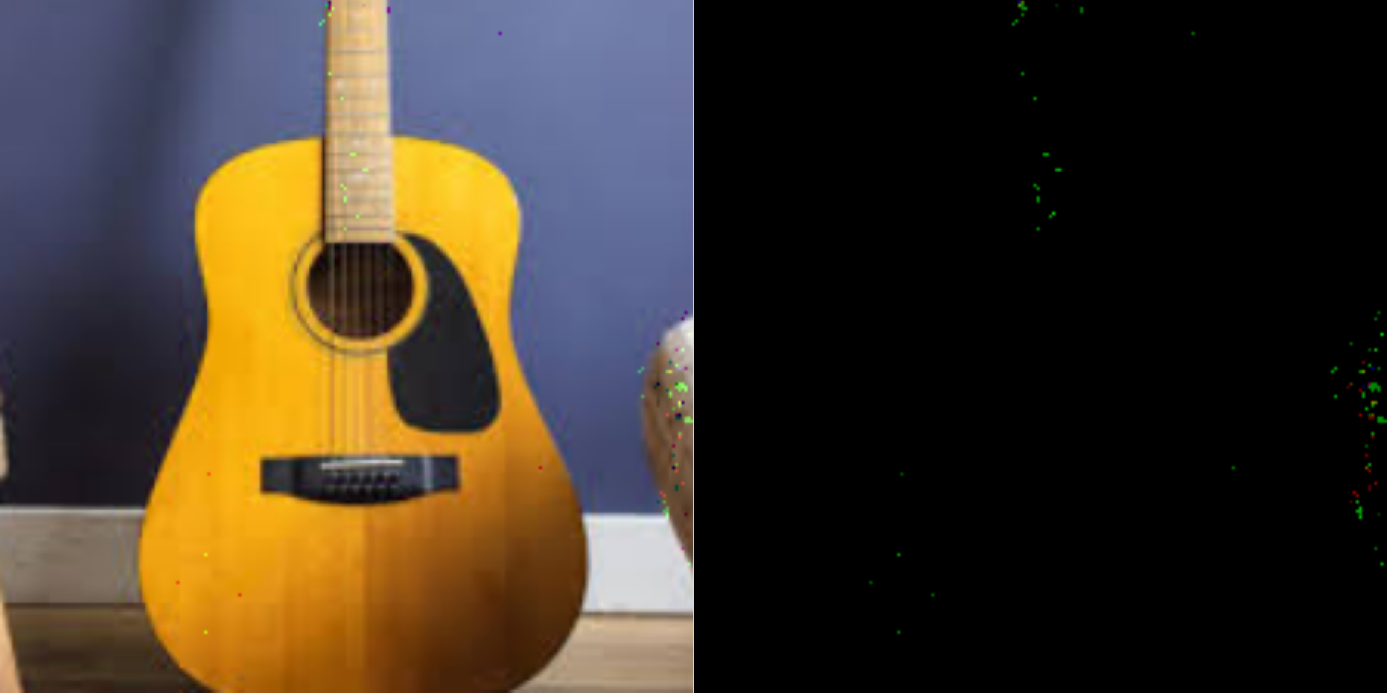}
                \caption{Ours \\ 
                ``tailed frog",~ \\
                    $ \norm{ \bm{\delta}}_{0} = 218$  }
                \label{fig:reb-ours-guitar}
            \end{subfigure}
            %
            % \vspace{-0.5 em}
            \caption{
                Illustration of targeted attacks by sparse attack counterparts~(Top) and our approach~(Bottom).
                From left to right, the ground-truth classes are ``castle'', ``airship", 
                and ``guitar", respectively.
                %                %
                The incorrect predictions and the sparsity
                (\ie\ $\norm{ \bm{\delta}}_{0}$) are listed under each image.
            }
            \label{fig:exp-sparsity-targeted}
            % \vspace{-2.0 em}
\end{figure}

Next, we conduct qualitative experiments to evaluate the sparsity under the targeted attack scenario.
Figure~\ref{fig:exp-sparsity-targeted} shows the visualized results,
where our approach modifies only $123, 107, $ and $218$ pixels, respectively,
with its adversarial examples able to 
mislead the classifier into targeted mispredictions.
In contrast, C\&W $L_{0}$, GreedyFool, and Homotopy have to 
perturb  $3555$, $1070$, and $1731$ pixels, respectively,  to perform effective targeted attacks. %,
%significantly underperforming our approach.
%
These qualitative results demonstrate the effectiveness of our approach on sparsity.
We also conduct qualitative comparison under the non-targeted attack scenario,
with the results deferred to Appendix C.1 of supplementary materials.

\subsection{Interpreting the Adversarial Perturbation}
\label{sec:misleading}

% noise 
\begin{figure}[!t]
    \captionsetup[subfigure]{justification=centering}
    \centering
    \begin{subfigure}[t]{0.2\textwidth}
        \centering
         \includegraphics[width=\textwidth]{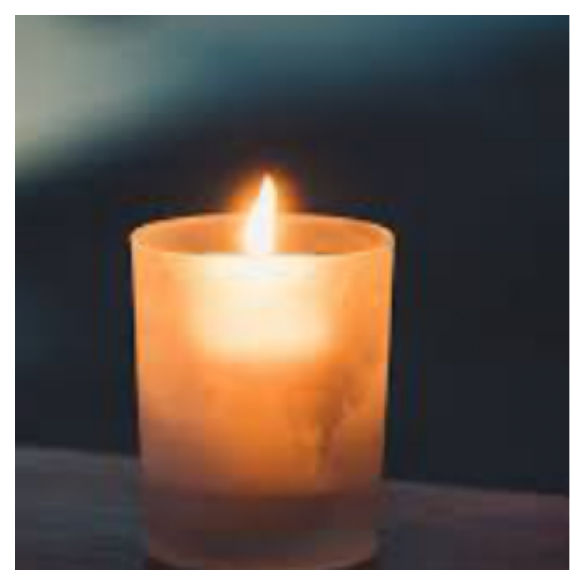}
        \caption{Clean image \\ ``candle''}
        \label{fig:exp-noise-clean}
    \end{subfigure}
    \centering
     \begin{subfigure}[t]{0.2\textwidth}
        \centering
         \includegraphics[width=\textwidth]{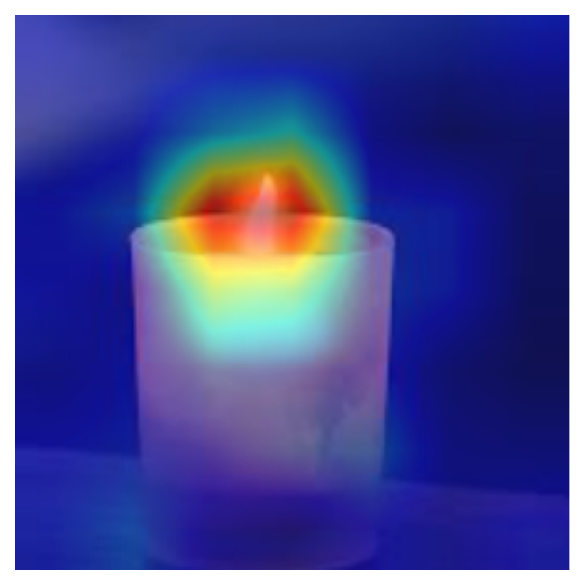}
        \caption{Grad-CAM \\ ``candle''}
        \label{fig:exp-noise-grad-cam-clean}
    \end{subfigure}
    \centering
    \begin{subfigure}[t]{0.2\textwidth}
        \centering
         \includegraphics[width=\textwidth]{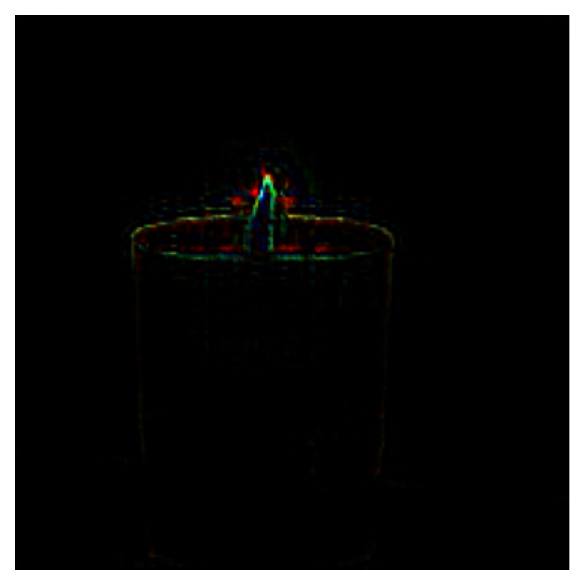}
        \caption{Guided Grad-CAM \\ ``candle''}
        \label{fig:exp-noise-guided-grad-cam}
    \end{subfigure}

    \centering
    \begin{subfigure}[t]{0.2\textwidth}
        \centering
         \includegraphics[width=\textwidth]{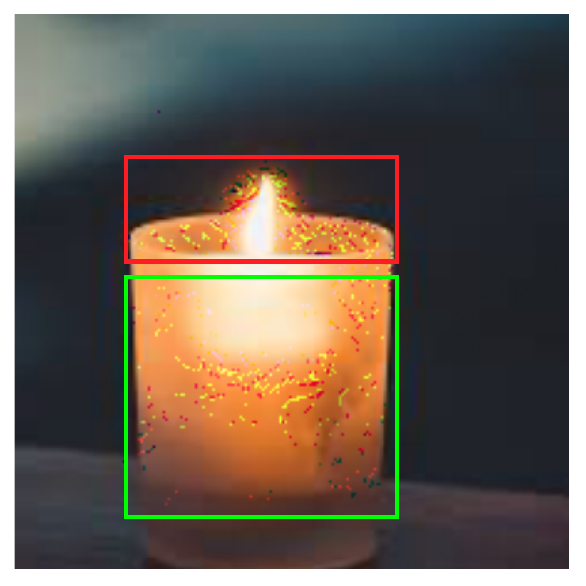}
        \caption{Adversarial example \\ ``toilet tissue''}
        \label{fig:exp-noise-adv}
    \end{subfigure}
     \begin{subfigure}[t]{0.2\textwidth}
        \centering
         \includegraphics[width=\textwidth]{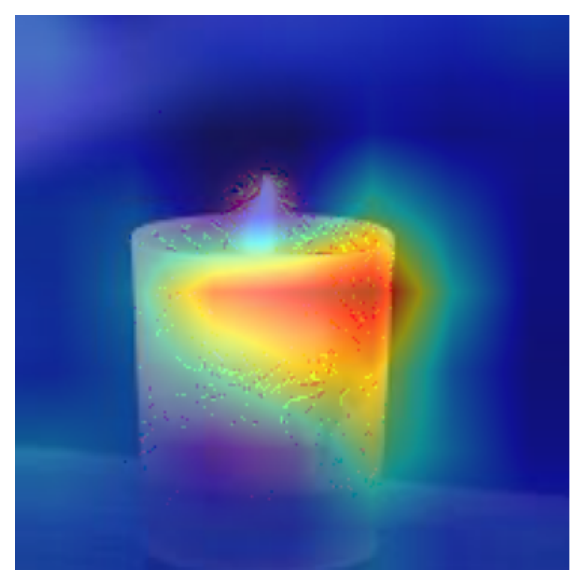}
        \caption{Grad-CAM \\ ``toilet tissue''}
        \label{fig:exp-noise-grad-cam-adv}
    \end{subfigure}
    \begin{subfigure}[t]{0.2\textwidth}
        \centering
         \includegraphics[width=\textwidth]{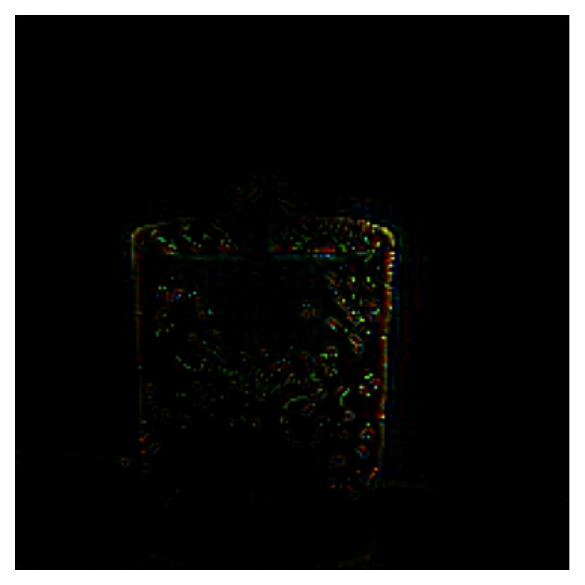}
        \caption{Guided Grad-CAM \\ ``toilet tissue''}
        \label{fig:exp-noise-guided-grad-cam-adv}
    \end{subfigure}
    % \vspace{-0.5em}
    \caption{
        Illustration of how adversarial perturbations computed by our approach
        mislead VGG-16 to predict the label of ``candle'' to ``toilet tissue''.
        \textbf{Left:} clean image and adversarial example with their predicted labels.
        \textbf{Middle:} Grad-CAM visualization.
        \textbf{Right:} Guided Grad-CAM visualization.
        }
    % \vspace{-1.5 em}
    \label{fig:exp-noise}
\end{figure}

Despite extensive attention to sparse attacks,
limited work has ever clearly explained how adversarial perturbations 
mislead DNNs into incorrect predictions.
This is due to the mediocre performance results of prior sparse attacks on sparsity, 
as discussed in Section~\ref{sec:sparsity}.
We aim to fill this gap by unveiling the mystery underlying the proposed adversarial perturbation,
resorting to Grad-CAM and Guided Grad-CAM visualizations~\cite{selvaraju:iccv17:grad-cam}.
Specifically, we let the prediction made by a well-trained VGG-16 
as the decision of interest for Grad-CAM and Guided Grad-CAM visualizations.
As such, we can interpret adversarial perturbations by comparing visualizations
on clean images with those on their corresponding adversarial examples.

Figure~\ref{fig:exp-noise} exhibits the experimental results,
where our generated perturbations mislead VGG-16 to 
mispredict ``toilet tissue'' on an image as ``candle''.
In particular, Figures~\ref{fig:exp-noise-clean}, \ref{fig:exp-noise-grad-cam-clean} and \ref{fig:exp-noise-guided-grad-cam} show visualizations on the clean image.
We observed that the ``candle wick'' is the critical region for making a correct prediction on the ``candle''.
By contrast, Figures~\ref{fig:exp-noise-adv}, \ref{fig:exp-noise-grad-cam-adv} and \ref{fig:exp-noise-guided-grad-cam-adv}
depict visualizations of the adversarial example generated by our approach. 
From Figure~\ref{fig:exp-noise-adv},
we observe that the adversarial perturbation
can be divided into two categories,
namely, ``obscuring noise" (\ie\ perturbations within the red box) 
and ``leading noise" (\ie\ perturbations within the green box).
The former prevents the classifier from identifying the true label
by covering essential features of the true class, 
\ie\ obscuring the feature of ``candle wick'',
whereas the latter leads the classifier to mispredict the adversarial example
by adding the essential features of the targeted class,
\ie\ adding noise within the shape of ``toilet tissue'';
see Figures~\ref{fig:exp-noise-grad-cam-clean} versus \ref{fig:exp-noise-grad-cam-adv}
and Figures~\ref{fig:exp-noise-guided-grad-cam} versus \ref{fig:exp-noise-guided-grad-cam-adv}.

We also conduct experiments to show how the two types of perturbations
mislead ResNet-50 to make an incorrect prediction as ``wing'' on an image of ``canoe'',
with the results deferred to Appendix C.2 of the supplementary materials.
To the best of our knowledge, 
we are the first to interpret how adversarial perturbations mislead DNNs into incorrect predictions, by discovering the ``obscuring noise'' and ``leading noise''.

\subsection{Attacking Robust Models}
\label{sec:strength}
%
%
%
% attack strength
% \input{C-1-table-strength.tex}
%
\begin{table} [!t]  % [htbp] %
   % \small
    \scriptsize
    \centering
    \setlength\tabcolsep{8 pt}
   \caption{The fooling rates ($\%$) on different robust models 
   trained adversarially by PGD-AT and Fast-AT, respectively,
   with the best results are shown in bold}
   % \vspace{-0.5 em}
   \begin{tabular}{@{}|c|cc|ccc|@{}}
    \toprule
    \multirow{2}{*}{Method}         & \multicolumn{2}{c|}{PGD-AT}                                           & \multicolumn{3}{c|}{Fast-AT}                                                                                  \\ \cmidrule(l){2-6} 
                                    & \multicolumn{1}{c}{MNIST}                & CIFAR-10                  & \multicolumn{1}{c}{MNIST}                    & \multicolumn{1}{c}{CIFAR-10}                 & ImageNet     \\ \midrule
    C\&W $L_{0}$                    & \multicolumn{1}{c}{$11.8$}               & $52.8$                    & \multicolumn{1}{c}{$12.1$}                  & \multicolumn{1}{c}{$62.3$}                    & $67.8$       \\  
    GreedyFool                        & \multicolumn{1}{c}{$3.6$}                & $26.8$                    & \multicolumn{1}{c}{$4.9$}                    & \multicolumn{1}{c}{$23.5$}                   & $24.9$       \\ 
    FMN                       & \multicolumn{1}{c}{$7.9$}                & $55.0$                    & \multicolumn{1}{c}{$9.8$}                    & \multicolumn{1}{c}{$52.9$}                   & $54.1$       \\ 
    Homotopy                      & \multicolumn{1}{c}{$8.5$}                & $51.9$                    & \multicolumn{1}{c}{$7.2$}                    & \multicolumn{1}{c}{$53.2$}                   & $58.7$       \\ 
    Sparse-RS                     & \multicolumn{1}{c}{$10.3$}                & $54.6$                    & \multicolumn{1}{c}{$10.9$}                    & \multicolumn{1}{c}{$57.1$}                   & $56.3$       \\ 
    SA-MOO                       & \multicolumn{1}{c}{$11.5$}               & $56.9$                    & \multicolumn{1}{c}{$7.8$}                    & \multicolumn{1}{c}{$58.9$}                   & $62.8$       \\
    EGS-TSSA                        & \multicolumn{1}{c}{$10.7$}               & $57.8$                    & \multicolumn{1}{c}{$8.4$}                    & \multicolumn{1}{c}{$61.6$}                   & $63.8$       \\ 
    BruSLeAttack                       & \multicolumn{1}{c}{$11.2$}               & $61.2$                    & \multicolumn{1}{c}{$9.6$}                    & \multicolumn{1}{c}{$58.9$}                   & $65.1$       \\ \midrule
    \textbf{Ours}                  & \multicolumn{1}{c}{$\mathbf{13.8}$}      & $\mathbf{62.5}$            & \multicolumn{1}{c}{$\mathbf{14.6}$}          & \multicolumn{1}{c}{$\mathbf{66.4}$}          & $\mathbf{72.1}$      \\ \bottomrule
    \end{tabular}
   \label{tab:strength}
\end{table}
We next compare our approach to sparse attacks in terms of attack strength,
\ie\ the fooling rate on robust models trained adversarially by 
PGD Adversarial Training (PGD-AT)~\cite{madry2018towards}
or Fast Adversarial Training (Fast-AT)~\cite{wong2020fast}.
We set $\epsilon=0.3$, $\epsilon=8/255$, $\epsilon=2/255$ for MNIST, CIFAR-10, and ImageNet, respectively.
Table~\ref{tab:strength} presents the experimental results.

From Table~\ref{tab:strength}, we observe that our approach achieves
the most powerful attacks under all scenarios, 
with the best fooling rate of $14.6 \%$, $66.4 \%$, and $72.1 \%$ on MNIST,
CIFAR-10, and ImageNet, respectively.
The reason is that 
\romsm{1}) our novel loss function (\ie\ \eqref{eq:final-loss})
significantly augments the adversary property;
and \romsm{2}) the proposed theoretically sound box constraint strategy
yields valid adversarial examples without the need of clipping invalid pixels,
empowering our approach to generate more optimized solutions.
Meanwhile, all sparse attacks, including our approach,
achieve smaller fooling rates on MNIST than those on CIFAR-10 and ImageNet.
This is because adversarial training can significantly 
improve the model's robustness on the simple dataset.
But, for CIFAR-10 (or ImageNet), our approach can achieve the  averaged fooling rate of $62.5 \%$ and of $66.4 \%$ (or $72.1 \%$), respectively,  
on PGD-AT and Fast-AT.

% transferability
%\input{C-1-table-trans.tex}
%
%
\begin{table} [!t]
    % \small
    \scriptsize
    \centering
    \setlength\tabcolsep{5 pt}
    \caption{Comparison of transferability (\ie\ Mean Fooling Rate ($\%$)) under ImageNet.
        The first column  denotes the models where adversarial samples  are generated while  the first row indicates the target models for attacking,
        with the best results shown in bold
        and $\ast$ indicating white-box attacks
        %
        % The transferability rates of I-FGSM are listed in the 8th and 15th rows.
        }
    % \vspace{-0.5 em}
    \begin{tabular}{@{}|c|c|c|c|c|c|@{}}
        \toprule
        Model                                   & Method            & \textbf{VGG-16}               & \textbf{VGG-19}               & \textbf{ResNet-101}       & \textbf{ResNet-152}   \\ \midrule
        \multirow{6}{*}{\textbf{VGG-19}}        & C\&W $L_{0}$      & $84.6$                        & $96.2^\ast$                   & $55.6$                    & $47.5$                 \\ 
                                                & GreedyFool         & $10.6$                        & $19.2^\ast$                   & $8.2$                     & $9.0$                 \\ 
                                                & FMN        & $83.4$                        & $91.8^\ast$                   & $43.0$                    & $46.8$                \\ 
                                                & Homotopy         & $65.8$                        & $87.6^\ast$                   & $24.0$                    & $9.8$                 \\ 
                                                & Sparse-RS               & $85.2$                        & $96.0^\ast$                   & $46.8$                    & $48.2$                 \\ 
                                                & SA-MOO          & $86.2$                        & $92.8^\ast$                   & $44.5$                    & $49.2$                \\ 
                                                & EGS-TSSA         & $87.7$                        & $93.2^\ast$                   & $51.3$                    & $51.2$                \\ 
                                                & BruSLeAttack            & $89.4$                        & $94.4^\ast$                   & $56.7$                    & $51.8$                \\  \cmidrule(l){2-6} 
                                                & \textbf{Ours}     & $\mathbf{99.1}$               & $\mathbf{99.9^\ast}$          & $\mathbf{63.3}$           & $\mathbf{55.1}$       \\ \midrule
                                                %& I-FGSM            & $99.5$                        & $99.9^\ast$                   & $76.7$                    & $72.5$                \\ \midrule
        \multirow{6}{*}{\textbf{ ResNet-101}}   & C\&W $L_{0}$      & $82.6$                        & $81.9$                        & $91.2^\ast$                & $50.5$                 \\
                                                & GreedyFool          & $11.4$                        & $13.4$                        & $9.4^\ast$                 & $8.4$                 \\ 
                                                & FMN         & $82.6$                        & $81.8$                        & $60.1^\ast$                & $41.6$                \\ 
                                                & Homotopy        & $63.6$                        & $63.6$                        & $72.0^\ast$                & $46.2$                \\ 
                                                & Sparse-RS              & $85.2$                        & $76.8$                        & $89.4^\ast$                & $65.6$                \\ 
                                                & SA-MOO          & $86.6$                        & $84.3$                        & $91.1^\ast$                & $61.3$                \\ 
                                                & EGS-TSSA        & $86.1$                        & $83.8$                        & $91.5^\ast$                & $62.6$                \\ 
                                                & BruSLeAttack            & $86.8$                        & $84.2$                        & $92.6^\ast$                & $67.8$                \\\cmidrule(l){2-6} 
                                                & \textbf{Ours}     & $\mathbf{87.6}$               & $\mathbf{86.3}$               & $\mathbf{99.9^\ast}$      & $\mathbf{89.9}$       \\ \bottomrule
                                                %& I-FGSM            & $89.5$                        & $88.2$                        & $99.8^\ast$               & $91.6$                \\ 
        \end{tabular}
    \label{tab:exp-trans}
    % \vspace{-2.0 em}
\end{table}

\subsection{Comparison on Transferability}
\label{sec:trans}

We take two  models,  \ie\ VGG-19 and ResNet-101, 
to generate adversarial examples, 
which are then employed to attack four different classifiers, \ie\ VGG-16, VGG-19, ResNet-101, and ResNet-152.
Table~\ref{tab:exp-trans} presents the comparative results of our approach and sparse attack counterparts
%, \ie\ C\&W $L_{0}$, OnePixel, SparseFool, GreedyFool, FMN, and Homotopy,  
in terms of transferability. 
Notably, the adversarial examples generated via VGG-19 (or ResNet-101) to attack VGG-19 (or ResNet-101) are considered as white-box attacks, while all others are black-box attacks.
We observe that compared to other sparse attacks, our approach achieves the best performance under all scenarios,
with the largest fooling rate of $99.1\%$ (or $99.9 \%$) under the black-box (or white-box) attack.
This demonstrates that modifying a few pixels is sufficient to make our approach enjoy high transferability.
%

% complexity
% \input{C-1-table-complexity.tex}
%
%
\begin{table} [!t] %[htbp] %
    % \tiny
    \scriptsize
    \centering
    \setlength\tabcolsep{5 pt}
   \caption{Comparing various sparse attacks in terms of computational complexity under different models,
   with the best results shown in bold
    }
   % \vspace{-0.5 em}
   \begin{tabular}{@{}|c|cccc|@{}}
    \toprule
    \multirow{2}{*}{Method}     & \multicolumn{4}{c|}{Time Cost~(s)}                                                                              \\ \cmidrule(l){2-5} 
                                & \multicolumn{1}{c|}{VGG-16}           & \multicolumn{1}{c|}{VGG-19}           & \multicolumn{1}{c|}{ResNet-101}           & ResNet-152    \\ \midrule
    C\&W $L_{0}$                & \multicolumn{1}{c|}{$20.8$}           & \multicolumn{1}{c|}{$23.4$}           & \multicolumn{1}{c|}{$28.7$}               & $43.6$        \\ 
    Homotopy                    & \multicolumn{1}{c|}{$519.4$}          & \multicolumn{1}{c|}{$531.5$}          & \multicolumn{1}{c|}{$1011.9$}             & $1462.2$      \\
    Sparse-RS                   & \multicolumn{1}{c|}{$69.0$}           & \multicolumn{1}{c|}{$76.4$}           & \multicolumn{1}{c|}{$71.6$}               & $99.7$        \\ 
    SA-MOO                  & \multicolumn{1}{c|}{$537.4$}          & \multicolumn{1}{c|}{$628.1$}          & \multicolumn{1}{c|}{$1123.8$}             & $1379.1$      \\ 
    EGS-TSSA                  & \multicolumn{1}{c|}{$71.8$}           & \multicolumn{1}{c|}{$79.5$}           & \multicolumn{1}{c|}{$120.9$}              & $175.9$       \\ 
    BruSLeAttack                  & \multicolumn{1}{c|}{$87.2$}           & \multicolumn{1}{c|}{$93.7$}           & \multicolumn{1}{c|}{$142.2$}              & $212.3$       \\  \midrule
    \textbf{Ours}               & \multicolumn{1}{c|}{$\bm {4.6}$}     & \multicolumn{1}{c|}{$\bm{5.3}$}      & \multicolumn{1}{c|}{$\bm{18.7}$}         & $\bm{27.5}$    \\ \bottomrule
    %I-FGSM                      & \multicolumn{1}{c|}{$0.16$}           & \multicolumn{1}{c|}{$0.18$}           & \multicolumn{1}{c|}{$0.37$}               & $0.57$       \\ \bottomrule
   \end{tabular}
   \label{tab:complexity}
   % \vspace{-1.0 em}
\end{table}

%\vspace{-0.5em}
\subsection{Comparison on Computational Complexity}
\label{sec:complexity}
We conduct the white-box attacks under ImageNet 
to show the superior computational efficiency of our approach for high-dimensional data. 
Four classifiers, \ie\ VGG-16, VGG-19, ResNet-101, and ResNet-152, are taken into account.
Table~\ref{tab:complexity} lists the experimental results.
Clearly, our approach runs much faster than all sparse attack counterparts (\ie\ C\&W $L_{0}$, Homotopy, Sparse-RS, SA-MOO, EGS-TSSA, and BruSLeAttack) for all examined classifiers, 
enjoying $3.0$x, $62.8$x, $4.6$x, $64.0$x, $8.0$x, and $13.0$x, and 
computational speedups on average.
Specifically, our approach takes only  $4.6s$, $5.3s$, $18.7s$, and $27.5s$, 
on VGG-16, VGG-19, ResNet-101, and ResNet-152, respectively.
This is because our reparameterization technique (\ie\ \eqref{eq:derivetive})
can efficiently approximate the NP-hard $l_{0}$ optimization problem
and hence substantially accelerate the convergence.

We also conducted ablation studies to exhibit the hyperparameter sensitivity
of $a$, $\lambda$, and $\tau$ 
respectively in \eqref{eq:derivetive}, \eqref{eq:convergence}, and \eqref{eq:final-loss},
with their details deferred to Section C.3 of supplementary materials.

\section{Conclusion}
\label{sec:conslusion}
This paper has addressed a sparse attack 
that yields interpretable adversarial examples,
thanks to their superb sparsity.
Meanwhile, our approach enjoys fast convergence, high transferability, and powerful attack strength.
The key idea is to approximate the NP-hard $l_{0}$ optimization problem
via a theoretical sound reparameterization technique,
making direct optimization of sparse perturbations computationally tractable.
Besides, a novel loss function and a theoretically sound box constraint strategy 
have been proposed to make our solution generate superior adversarial examples, 
yielding fast, transferable, and powerful sparse adversarial attacks. 
Experimental results demonstrate that our approach clearly outperforms \sota\ sparse attacks 
in terms of computational efficiency, transferability, and attack intensity.
In addition,
theoretical and empirical results verify that our approach yields sparser adversarial  examples,
empowering us to experimentally interpret how  adversarial examples mislead \sota\ DNNs
into incorrect predictions. 
Our work is expected to 
\romsm{1}) serve as the benchmark for evaluating the robustness of DNNs,
and \romsm{2}) shed light on future work about interpreting the vulnerability of DNNs.

\section*{Acknowledgments}
This work was supported in part by NSF under Grants  2019511, 2315613, 2325564, 2438898, 2523997, 2534286, 2315612, and 2332638.
Any opinions and findings expressed in the paper are those of the authors and do not necessarily reflect the views of funding agencies.

\bibliographystyle{plain}
\bibliography{main}

\begin{thebibliography}{10}

\bibitem{ali2025enabling}
Ahsan Ali, Xiaolong Ma, Syed Zawad, Paarijaat Aditya, Istemi~Ekin Akkus, Ruichuan Chen, Lei Yang, and Feng Yan.
\newblock Enabling scalable and adaptive machine learning training via serverless computing on public cloud.
\newblock {\em Performance Evaluation}, 167:102451, 2025.

\bibitem{carlini2017towards}
Nicholas Carlini and David~A. Wagner.
\newblock Towards evaluating the robustness of neural networks.
\newblock In {\em {IEEE} Symposium on Security and Privacy}, 2017.

\bibitem{chen20hopskip}
Jianbo Chen, Michael~I. Jordan, and Martin~J. Wainwright.
\newblock Hopskipjumpattack: {A} query-efficient decision-based attack.
\newblock In {\em {IEEE} Symposium on Security and Privacy, 2020}, 2020.

\bibitem{lin20:iclr20:more_data}
Lin Chen, Yifei Min, Mingrui Zhang, and Amin Karbasi.
\newblock More data can expand the generalization gap between adversarially robust and standard models.
\newblock In {\em International Conference on Machine Learning (ICML)}, 2020.

\bibitem{tiankuo:iclr25:bonemet}
Tiankuo Chu, Fudong Lin, Shubo Wang, Jason Jiang, Wiley~Jia{-}Wei Gong, Xu~Yuan, and Liyun Wang.
\newblock Bonemet: An open large-scale multi-modal murine dataset for breast cancer bone metastasis diagnosis and prognosis.
\newblock In {\em The Thirteenth International Conference on Learning Representations (ICLR)}, 2025.

\bibitem{croce:aaai22:sparse_rs}
Francesco Croce, Maksym Andriushchenko, Naman~D. Singh, Nicolas Flammarion, and Matthias Hein.
\newblock Sparse-rs: {A} versatile framework for query-efficient sparse black-box adversarial attacks.
\newblock In {\em AAAI}, 2022.

\bibitem{croce19sparse}
Francesco Croce and Matthias Hein.
\newblock Sparse and imperceivable adversarial attacks.
\newblock In {\em ICCV}, 2019.

\bibitem{devlin2019bert}
Jacob Devlin, Ming{-}Wei Chang, Kenton Lee, and Kristina Toutanova.
\newblock {BERT:} pre-training of deep bidirectional transformers for language understanding.
\newblock In {\em Conference of the North American Chapter of the Association for Computational Linguistics: Human Language Technologies (NAACL-HLT)}, 2019.

\bibitem{dong2020greedyfool}
Xiaoyi Dong, Dongdong Chen, Jianmin Bao, Chuan Qin, Lu~Yuan, Weiming Zhang, Nenghai Yu, and Dong Chen.
\newblock Greedyfool: Distortion-aware sparse adversarial attack.
\newblock In {\em NeurIPS}, 2020.

\bibitem{dong2018boosting}
Yinpeng Dong, Fangzhou Liao, Tianyu Pang, Hang Su, Jun Zhu, Xiaolin Hu, and Jianguo Li.
\newblock Boosting adversarial attacks with momentum.
\newblock In {\em Conference on Computer Vision and Pattern Recognition (CVPR)}, 2018.

\bibitem{dong2019TI-FGSM}
Yinpeng Dong, Tianyu Pang, Hang Su, and Jun Zhu.
\newblock Evading defenses to transferable adversarial examples by translation-invariant attacks.
\newblock In {\em Conference on Computer Vision and Pattern Recognition (CVPR)}, 2019.

\bibitem{dosovitskiy:iclr21:vit}
Alexey Dosovitskiy, Lucas Beyer, Alexander Kolesnikov, Dirk Weissenborn, Xiaohua Zhai, Thomas Unterthiner, Mostafa Dehghani, Matthias Minderer, Georg Heigold, Sylvain Gelly, Jakob Uszkoreit, and Neil Houlsby.
\newblock An image is worth 16x16 words: Transformers for image recognition at scale.
\newblock In {\em ICLR}, 2021.

\bibitem{eykholt2018robust}
Kevin Eykholt, Ivan Evtimov, Earlence Fernandes, Bo~Li, Amir Rahmati, Chaowei Xiao, Atul Prakash, Tadayoshi Kohno, and Dawn Song.
\newblock Robust physical-world attacks on deep learning visual classification.
\newblock In {\em CVPR}, 2018.

\bibitem{fan20factor}
Yanbo Fan, Baoyuan Wu, Tuanhui Li, Yong Zhang, Mingyang Li, Zhifeng Li, and Yujiu Yang.
\newblock Sparse adversarial attack via perturbation factorization.
\newblock In {\em ECCV}, 2020.

\bibitem{goodfellow2015explaining}
Ian~J. Goodfellow, Jonathon Shlens, and Christian Szegedy.
\newblock Explaining and harnessing adversarial examples.
\newblock In {\em International Conference on Learning Representations (ICLR)}, 2015.

\bibitem{he2016deep}
Kaiming He, Xiangyu Zhang, Shaoqing Ren, and Jian Sun.
\newblock Deep residual learning for image recognition.
\newblock In {\em {IEEE} Conference on Computer Vision and Pattern Recognition (CVPR)}, 2016.

\bibitem{yi:ijcai21:oversampling}
Yi~He, Fudong Lin, Xu~Yuan, and Nian{-}Feng Tzeng.
\newblock Interpretable minority synthesis for imbalanced classification.
\newblock In {\em Proceedings of the Thirtieth International Joint Conference on Artificial Intelligence (IJCAI)}, pages 2542--2548, 2021.

\bibitem{jumper:nature21:alpha_fold}
John Jumper, Richard Evans, Alexander Pritzel, Tim Green, Michael Figurnov, Olaf Ronneberger, Kathryn Tunyasuvunakool, Russ Bates, Augustin {\v{Z}}{\'\i}dek, Anna Potapenko, et~al.
\newblock Highly accurate protein structure prediction with alphafold.
\newblock {\em Nature}, 596(7873):583--589, 2021.

\bibitem{kong2020physgan}
Zelun Kong, Junfeng Guo, Ang Li, and Cong Liu.
\newblock Physgan: Generating physical-world-resilient adversarial examples for autonomous driving.
\newblock In {\em {IEEE} Conference on Computer Vision and Pattern Recognition (CVPR)}, 2020.

\bibitem{krizhevsky2009cifar10}
Alex Krizhevsky, Geoffrey Hinton, et~al.
\newblock Learning multiple layers of features from tiny images.
\newblock {\em Technical Report, University of Toronto}, 2009.

\bibitem{bartlett2012imagenet}
Alex Krizhevsky, Ilya Sutskever, and Geoffrey~E. Hinton.
\newblock Imagenet classification with deep convolutional neural networks.
\newblock In {\em Advances in Neural Information Processing Systems (NIPS)}, 2012.

\bibitem{kurakin2017adversarial}
Alexey Kurakin, Ian~J. Goodfellow, and Samy Bengio.
\newblock Adversarial examples in the physical world.
\newblock In {\em International Conference on Learning Representations (ICLR)}, 2017.

\bibitem{kurakin2017scale}
Alexey Kurakin, Ian~J. Goodfellow, and Samy Bengio.
\newblock Adversarial machine learning at scale.
\newblock In {\em 5th International Conference on Learning Representations (ICLR)}, 2017.

\bibitem{lan2025contextual}
Guangchen Lan, Huseyin~A. Inan, Sahar Abdelnabi, Janardhan Kulkarni, Lukas Wutschitz, Reza Shokri, Christopher~G. Brinton, and Robert Sim.
\newblock Contextual integrity in {LLMs} via reasoning and reinforcement learning.
\newblock {\em arXiv preprint arXiv:2506.04245}, 2025.

\bibitem{lecun2010mnist}
Yann LeCun and Corinna Cortes.
\newblock {MNIST} handwritten digit database.
\newblock 2010.

\bibitem{fudong:iccv23:mmst_vit}
Fudong Lin, Summer Crawford, Kaleb Guillot, Yihe Zhang, Yan Chen, Xu~Yuan, et~al.
\newblock Mmst-vit: Climate change-aware crop yield prediction via multi-modal spatial-temporal vision transformer.
\newblock In {\em {IEEE/CVF} International Conference on Computer Vision (ICCV)}, pages 5751--5761, 2023.

\bibitem{fudong:cikm24:mat}
Fudong Lin, Jiadong Lou, Xu~Yuan, and Nian-Feng Tzeng.
\newblock Towards robust vision transformer via masked adaptive ensemble.
\newblock In {\em Proceedings of the 33rd ACM International Conference on Information and Knowledge Management (CIKM)}, pages 1389--1399, 2024.

\bibitem{fudong:ecml23:storm}
Fudong Lin, Xu~Yuan, Yihe Zhang, Purushottam Sigdel, Li~Chen, Lu~Peng, and Nian-Feng Tzeng.
\newblock Comprehensive transformer-based model architecture for real-world storm prediction.
\newblock In {\em Joint European Conference on Machine Learning and Knowledge Discovery in Databases (ECML-PKDD)}, pages 54--71, 2023.

\bibitem{liu2017delving}
Yanpei Liu, Xinyun Chen, Chang Liu, and Dawn Song.
\newblock Delving into transferable adversarial examples and black-box attacks.
\newblock In {\em ICLR}, 2017.

\bibitem{louizos:iclr18:l0-regularization}
Christos Louizos, Max Welling, and Diederik~P Kingma.
\newblock Learning sparse neural networks through l\_0 regularization.
\newblock In {\em ICLR}, 2018.

\bibitem{ma2024malletrain}
Xiaolong Ma, Feng Yan, Lei Yang, Ian Foster, Michael~E Papka, Zhengchun Liu, and Rajkumar Kettimuthu.
\newblock Malletrain: Deep neural networks training on unfillable supercomputer nodes.
\newblock In {\em Proceedings of the 15th ACM/SPEC International Conference on Performance Engineering}, pages 190--200, 2024.

\bibitem{madry2018towards}
Aleksander Madry, Aleksandar Makelov, Ludwig Schmidt, Dimitris Tsipras, and Adrian Vladu.
\newblock Towards deep learning models resistant to adversarial attacks.
\newblock In {\em International Conference on Learning Representations (ICLR)}, 2018.

\bibitem{min22:uai:curious}
Yifei Min, Lin Chen, and Amin Karbasi.
\newblock The curious case of adversarially robust models: More data can help, double descend, or hurt generalization.
\newblock In {\em Uncertainty in Artificial Intelligence}, 2021.

\bibitem{ming:cvpr24:egs_tssa}
Di~Ming, Peng Ren, Yunlong Wang, and Xin Feng.
\newblock Transferable structural sparse adversarial attack via exact group sparsity training.
\newblock In {\em Computer Vision and Pattern Recognition (CVPR)}, pages 24696--24705, 2024.

\bibitem{modas2019sparse}
Apostolos Modas, Seyed{-}Mohsen Moosavi{-}Dezfooli, and Pascal Frossard.
\newblock Sparsefool: {A} few pixels make a big difference.
\newblock In {\em {IEEE} Conference on Computer Vision and Pattern Recognition (CVPR)}, 2019.

\bibitem{mohsen2016deepfool}
Seyed{-}Mohsen Moosavi{-}Dezfooli, Alhussein Fawzi, and Pascal Frossard.
\newblock Deepfool: {A} simple and accurate method to fool deep neural networks.
\newblock In {\em Conference on Computer Vision and Pattern Recognition (CVPR)}, 2016.

\bibitem{papernot2016JSMA}
Nicolas Papernot, Patrick~D. McDaniel, Somesh Jha, Matt Fredrikson, Z.~Berkay Celik, and Ananthram Swami.
\newblock The limitations of deep learning in adversarial settings.
\newblock In {\em European Symposium on Security and Privacy (EuroS{\&}P)}, 2016.

\bibitem{maura:nips21:fmn}
Maura Pintor, Fabio Roli, Wieland Brendel, and Battista Biggio.
\newblock Fast minimum-norm adversarial attacks through adaptive norm constraints.
\newblock In {\em Neural Information Processing Systems (NeurIPS)}, pages 20052--20062, 2021.

\bibitem{rives:pans21:ESM-1b}
Alexander Rives, Joshua Meier, Tom Sercu, Siddharth Goyal, Zeming Lin, Jason Liu, Demi Guo, Myle Ott, C~Lawrence Zitnick, Jerry Ma, et~al.
\newblock Biological structure and function emerge from scaling unsupervised learning to 250 million protein sequences.
\newblock {\em Proceedings of the National Academy of Sciences}, 118(15), 2021.

\bibitem{russakovsky2015imagenet}
Olga Russakovsky, Jia Deng, Hao Su, Jonathan Krause, Sanjeev Satheesh, Sean Ma, Zhiheng Huang, Andrej Karpathy, Aditya Khosla, Michael Bernstein, Alexander~C. Berg, and Li~Fei-Fei.
\newblock {ImageNet Large Scale Visual Recognition Challenge}.
\newblock {\em International Journal of Computer Vision (IJCV)}, 2015.

\bibitem{selvaraju:iccv17:grad-cam}
Ramprasaath~R. Selvaraju, Michael Cogswell, Abhishek Das, Ramakrishna Vedantam, Devi Parikh, and Dhruv Batra.
\newblock Grad-cam: Visual explanations from deep networks via gradient-based localization.
\newblock In {\em ICCV}, 2017.

\bibitem{simonyan2014very}
Karen Simonyan and Andrew Zisserman.
\newblock Very deep convolutional networks for large-scale image recognition.
\newblock In {\em International Conference on Learning Representations (ICLR)}, 2015.

\bibitem{su2019one}
Jiawei Su, Danilo~Vasconcellos Vargas, and Kouichi Sakurai.
\newblock One pixel attack for fooling deep neural networks.
\newblock {\em {IEEE} Trans. Evol. Comput.}, 2019.

\bibitem{suya2020hybrid}
Fnu Suya, Jianfeng Chi, David Evans, and Yuan Tian.
\newblock Hybrid batch attacks: Finding black-box adversarial examples with limited queries.
\newblock In {\em USENIX}, 2020.

\bibitem{szegedy2014intriguing}
Christian Szegedy, Wojciech Zaremba, Ilya Sutskever, Joan Bruna, Dumitru Erhan, Ian~J. Goodfellow, and Rob Fergus.
\newblock Intriguing properties of neural networks.
\newblock In {\em International Conference on Learning Representations (ICLR)}, 2014.

\bibitem{thys2019fooling}
Simen Thys, Wiebe~Van Ranst, and Toon Goedem{\'{e}}.
\newblock Fooling automated surveillance cameras: Adversarial patches to attack person detection.
\newblock In {\em {IEEE} Conference on Computer Vision and Pattern Recognition Workshops (CVPRW)}, 2019.

\bibitem{tramer2018ensemble}
Florian Tram{\`{e}}r, Alexey Kurakin, Nicolas Papernot, Ian~J. Goodfellow, Dan Boneh, and Patrick~D. McDaniel.
\newblock Ensemble adversarial training: Attacks and defenses.
\newblock In {\em International Conference on Learning Representations (ICLR)}, 2018.

\bibitem{vaswani2017attention}
Ashish Vaswani, Noam Shazeer, Niki Parmar, Jakob Uszkoreit, Llion Jones, Aidan~N. Gomez, Lukasz Kaiser, and Illia Polosukhin.
\newblock Attention is all you need.
\newblock In {\em Advances in Neural Information Processing Systems (NIPS)}, 2017.

\bibitem{vo:iclr24:brusle_attack}
Viet~Quoc Vo, Ehsan Abbasnejad, and Damith Ranasinghe.
\newblock Brusleattack: a query-efficient score- based black-box sparse adversarial attack.
\newblock In {\em International Conference on Learning Representations(ICLR)}, 2024.

\bibitem{wiki:step_function}
Wikipedia.
\newblock Heaviside step function, 2024.
\newblock [Online; accessed 01-January-2024].

\bibitem{williams:cvpr23:sa_moo}
Phoenix~Neale Williams and Ke~Li.
\newblock Black-box sparse adversarial attack via multi-objective optimisation.
\newblock In {\em Computer Vision and Pattern Recognition (CVPR)}, 2023.

\bibitem{wong2020fast}
Eric Wong, Leslie Rice, and J.~Zico Kolter.
\newblock Fast is better than free: Revisiting adversarial training.
\newblock In {\em International Conference on Learning Representations (ICLR)}, 2020.

\bibitem{xie2019DI-FGSM}
Cihang Xie, Zhishuai Zhang, Yuyin Zhou, Song Bai, Jianyu Wang, Zhou Ren, and Alan~L. Yuille.
\newblock Improving transferability of adversarial examples with input diversity.
\newblock In {\em {IEEE} Conference on Computer Vision and Pattern Recognition (CVPR)}, 2019.

\bibitem{xu19structured}
Kaidi Xu, Sijia Liu, Pu~Zhao, Pin{-}Yu Chen, Huan Zhang, Quanfu Fan, Deniz Erdogmus, Yanzhi Wang, and Xue Lin.
\newblock Structured adversarial attack: Towards general implementation and better interpretability.
\newblock In {\em ICLR}, 2019.

\bibitem{zawad2025fedcust}
Syed Zawad, Xiaolong Ma, Jun Yi, Cheng Li, Minjia Zhang, Lei Yang, Feng Yan, and Yuxiong He.
\newblock Fedcust: Offloading hyperparameter customization for federated learning.
\newblock {\em Performance Evaluation}, 167:102450, 2025.

\bibitem{zeiler2014visualize}
Matthew~D. Zeiler and Rob Fergus.
\newblock Visualizing and understanding convolutional networks.
\newblock In {\em European Conference on Computer Vision}, 2014.

\bibitem{zhu2021homotopy}
Mingkang Zhu, Tianlong Chen, and Zhangyang Wang.
\newblock Sparse and imperceptible adversarial attack via a homotopy algorithm.
\newblock In {\em ICML}, 2021.

\end{thebibliography}

\newpage
\appendix

\section*{Outline}
This document supplements the main paper in three aspects.
First, Section~\ref{sec:sup-tech} presents detailed proofs of two propositions stated in the main paper.
Second, Section~\ref{sec:sup-algorithm} lists the algorithm flow of our approach.
Third, Section~\ref{sec:sup-exp} provides supporting experimental results.

\section{Proofs of Theoretical Results} \label{sec:sup-tech}

% This section supports the main paper by providing the proof of 
% Theorem 1, Proposition 1, and Proposition 2.

This section supports the main paper by providing the proofs of Proposition 1 and Proposition 2.

\subsection{Proof of Proposition 1} 
\label{sec:sup-proof-sparsity}
Here, we state again Proposition 1 in the main paper:
\begin{prop} \label{sup_thm:sparsity}
    (Sparsity) 
    Given an initial perturbation $ \bm{\delta} \in [-\epsilon, \epsilon]^{d}$
    under some $l_{\infty}$ constraint $\epsilon~(\epsilon > 0) $,
    let $\bm{w} \in \mathbb{R} ^{d} $ be the weight matrix used in our study,
    and $\pi( \cdot)$ be the ReLU function, 
    in terms of the $l_{0}$ norm, we have:
    \begin{equation} \label{eq:sparsity}
       \norm{ \pi (\bm{w} - \frac{\tau}{\epsilon} ) \odot \bm{\delta}}_{0} 
       ~\leq~ \norm{  \pi (\bm{w}) \odot \bm{\delta} }_{0}
       ~\leq~ \norm{ \bm{w} \odot \bm{\delta} }_{0}.
    \end{equation}
\end{prop}
%
% proof
\begin{proof}
    Notice that $\bm{\delta}$ appears in all three terms,
    so we only need to proof $\norm{ \pi (\bm{w} - \frac{\tau}{\epsilon} ) }_{0} ~\leq~ \norm{ \pi (\bm{w} ) }_{0} ~\leq~ \norm{ \bm{w} }_{0}$.

    Given the weight matrix $\bm{w}$,
    we consider the following four scenarios:
    \begin{equation} \label{eq:weight-detail}
        \bm{w} = \begin{cases}
            ~ w_{i},  & \textrm{for} ~w_{i} > \frac{\tau}{\epsilon} ~\textrm{and}~ i = 1, 2, \dots, d_{1},      \\
            ~ w_{j},         & \textrm{for} ~0 < w_{j} \leq \frac{\tau}{\epsilon} ~\textrm{and}~ j = 1, 2, \dots, d_{2},   \\
            ~ 0,         & \textrm{for} ~w_{m} = 0 ~\textrm{and}~ m = 1, 2, \dots, d_{3},   \\
            ~ w_{n},         & \textrm{for} ~w_{n} < 0 ~\textrm{and}~ n = 1, 2, \dots, d_{4}.   \\
        \end{cases},
    \end{equation}
    where $d_{i}$ indicates the number of elements in the $i$-th scenario,
    and $d_{1} + d_{2} + d_{3} + d_{4} = d$.
    Accordingly, $\pi (\bm{w})$ can be expressed as follows,
    \begin{equation} \label{eq:weight-relu-detail}
        \begin{gathered}
            \pi (\bm{w}) = \begin{cases}
                ~ w_{i},  & \textrm{for} ~w_{i} > \frac{\tau}{\epsilon} ~\textrm{and}~ i = 1, 2, \dots, d_{1},      \\
                ~ w_{j},         & \textrm{for} ~0 < w_{j} \leq \frac{\tau}{\epsilon} ~\textrm{and}~ j = 1, 2, \dots, d_{2},   \\
                ~ 0,         & \textrm{for} ~w_{m} = 0 ~\textrm{and}~ m = 1, 2, \dots, d_{3},   \\
                ~ 0,         & \textrm{for} ~w_{n} < 0 ~\textrm{and}~ n = 1, 2, \dots, d_{4}.   \\
            \end{cases}.
        \end{gathered}
    \end{equation}
    Similarly, for $\pi (\bm{w} - \frac{\tau}{\epsilon} )$, we have,
    \begin{equation} \label{eq:weight-relu-ours-detail}
        \begin{gathered}
            \pi (\bm{w} - \frac{\tau}{\epsilon} ) = \begin{cases}
                ~ w_{i} - \frac{\tau}{\epsilon} ,  & \textrm{for} ~w_{i} > \frac{\tau}{\epsilon} ~\textrm{and}~ i = 1, 2, \dots, d_{1},      \\
                ~ 0,         & \textrm{for} ~0 < w_{j} \leq \frac{\tau}{\epsilon} ~\textrm{and}~ j = 1, 2, \dots, d_{2},   \\
                ~ 0,         & \textrm{for} ~w_{m} = 0 ~\textrm{and}~ m = 1, 2, \dots, d_{3},   \\
                ~ 0,         & \textrm{for} ~w_{n} < 0 ~\textrm{and}~ n = 1, 2, \dots, d_{4}.   \\
            \end{cases}.
        \end{gathered}
    \end{equation}

    Let $\mathds{1}_{S} (\cdot)$ be the sparsity indicator function 
    for indicating the non-zero elements in a matrix, expressed as below, 
    \begin{equation} \label{eq:indicator}
        \mathds{1}_{S} (x)= \begin{cases}
            ~1,  & \text{if}~x   \neq 0 \\
            ~0,         & \textrm{otherwise}
        \end{cases}.
    \end{equation}
    As such, the $l_{0}$ norm for the weight $\bm{w}$ can be re-writed as,
    \begin{equation} \label{eq:l0-weight}
        \begin{gathered}
            \norm{ \bm{w} }_{0} = \sum_{k=1}^{d} \mathds{1}_{S} (w_{k})
            = \sum_{i=1}^{d_{1}} \mathds{1}_{S} (w_{i}) + \sum_{j=1}^{d_{2}} \mathds{1}_{S} (w_{j}) 
            + \sum_{m=1}^{d_{3}} \mathds{1}_{S} (w_{m}) + \sum_{n=1}^{d_{4}} \mathds{1}_{S} (w_{n}) \\
            = d_{1} + d_{2} + d_{4}.
        \end{gathered}
    \end{equation}
    Similarly, for the $l_{0}$ norm of $ \pi (\bm{w}) $ and $ \pi (\bm{w} - \frac{\tau}{\epsilon} ) $, we have, 
    \begin{equation} \label{eq:l0-trick}
        \begin{gathered}
            \norm{ \pi (\bm{w}) }_{0} = \sum_{k=1}^{d} \mathds{1}_{S} \left( \pi (w_{k}) \right) = d_{1} + d_{2}, \\
            \norm{ \pi (\bm{w} - \frac{\tau}{\epsilon} ) }_{0} = \sum_{k=1}^{d} \mathds{1}_{S} \left( \pi (w_{k} - \frac{\tau}{\epsilon} ) \right) = d_{1}. \\
        \end{gathered}
    \end{equation}
    Since $d_{i} \geq 0$, we have $d_{1} \leq d_{1} + d_{2} \leq d_{1} + d_{2} + d_{4} $, 
    indicating that $\norm{ \pi (\bm{w} - \frac{\tau}{\epsilon} ) }_{0} ~\leq~ \norm{ \pi (\bm{w} ) }_{0} ~\leq~ \norm{ \bm{w} }_{0}$.
    This confirms that $\norm{ \pi (\bm{w} - \frac{\tau}{\epsilon} ) \odot \bm{\delta}}_{0} 
    ~\leq~ \norm{  \pi (\bm{w}) \odot \bm{\delta} }_{0}
    ~\leq~ \norm{ \bm{w} \odot \bm{\delta} }_{0}.$

\end{proof}

\subsection{Proof of Proposition 2} 
\label{sec:s-proof}

We state again Eq. (22) and Proposition 2 in the main paper:

\begin{equation} \label{eq:bound}
    \bm{w}^{\prime} = \bm{\Omega} \odot H \left( \pi (\bm{w} - \frac{\tau}{\epsilon} ) \right).
\end{equation}

\begin{prop} \label{sup_thm:bound}
    (Box constraint)
   Let $\bm{x} \in [0, 1]^\textrm{d}$ be a natural image, 
   $\bm{\delta} \in [-\epsilon, \epsilon]^{d}$ be the initial perturbation 
   %computed by I-FGSM 
   under a $l_{\infty}$ constraint $\epsilon$ ($\epsilon > 0$), 
   and $\bm{w}^{\prime}$ be the output of \eqref{eq:bound}.
   If ~$\bm{\Omega} = \min(\frac{\bm{x}}{\epsilon}, \frac{1-\bm{x}}{\epsilon})$,
   the resulting adversarial example is valid, \ie\ 
   \begin{equation} \label{eq:thm}
       \bm{x} + \bm{w}^{\prime} \odot \bm{\delta} \in [0, 1]^{\textrm{d}}.
   \end{equation}
\end{prop}

% proof
\begin{proof}
    First, note that $ \bm{x} $, $ \bm{\delta} $, $ \bm{w} $, and $ \bm{\Omega}$ have the same dimension,
    and $H (\cdot)$ and $\pi (\cdot)$ represent the Heaviside step and the  ReLU function, respectively.
    Let $g ( \bm{w} ) = H \left( \pi ( \bm{w} - \frac{\tau}{\epsilon} ) \right)$, 
    we can rewrite \eqref{eq:thm} as below:
    \begin{equation} \label{eq:gw}
        \bm{x} + \bm{\Omega} \odot g (\bm{w}) \odot \bm{\delta} \in [0, 1]^{\textrm{d}}.
    \end{equation}

    Since we have $ \forall w_{j} \in \bm{w}, ~g(w_{j}) = \{0, 1 \}$,
    we only need to consider two cases for $g (\bm{w})$. 
    Obivisely, the case when $g (w_{j}) = 0$, 
    we always have $x_{j} + \Omega_{j} \cdot 0 \cdot \delta_{j} = x_{j} \in [0, 1]$.

    Considering the another case when $g (w_{j}) = 1$, we only need to proof the following equation:
    \begin{equation} \label{eq:remove-gw}
        \bm{x} + \bm{\Omega}  \odot \bm{\delta} \in [0, 1]^{\textrm{d}}.
    \end{equation}

    Since the initial perturbation $ \bm{\delta}$ is computed 
    under a $l_{\infty}$ constraint $\epsilon$ ($\epsilon>0$), 
   corresponding to each pixel $j$, we have:
    \begin{equation} \label{eq:delta}
        \delta_{j} = \begin{cases}
            ~\epsilon,  & \nabla_{x_{j}}   > 0 \\
            ~0,         & \nabla_{x_{j}}   = 0 \\
            \text{-}\epsilon,   & \nabla_{x_{j}}   < 0 
        \end{cases},
    \end{equation}
    where $\nabla_{x_{j}}$ is the gradients in Eq. (3) %or Eq. (4)
    of the main paper, 
    \ie\ $\nabla_{x_{j}} J( \bm{x}, ~y_{\textrm{true}})$. %or $\nabla_{x_{j}} J( \bm{x}, ~y_{\textrm{adv}})$.

    Obviously, the case when $\delta_{j} = 0$,
    we always have $x_{j} + \Omega_{j} \cdot \delta_{j} \in [0, 1]$.

    Now, let $\Omega_{j} \geq 0$ for all pixels $j$.
    Considering the case when $\delta_{j} = \epsilon$.
    We always have $x_{j} + \Omega_{j} \cdot \epsilon \geq 0$,
    the valid pixel only needs to satisfy:
    
    \begin{equation} \label{eq:positive eps}
        \begin{gathered} 
            x_{j} + \Omega_{j} \cdot \epsilon ~\leq ~1 \\
            \Omega_{j} ~\leq~ \frac{1 - x_{j}}{\epsilon}.
        \end{gathered}
    \end{equation}

    Similarly, the case when $\delta_{j} = \text{-}\epsilon$,
    we always have  $x_{j} + \Omega_{j} \cdot (-\epsilon) \leq 1$;
    hence, the valid pixel just needs to satisfy: 
    \begin{equation} \label{eq:negative eps}
        \begin{gathered}            
            x_{j} + \Omega_{j} \cdot (\text{-}\epsilon) \geq 0 \\
            \Omega_{j} ~\leq~ \frac{x_{j}}{\epsilon}.
        \end{gathered}
    \end{equation}

    Combining \eqref{eq:positive eps} and \eqref{eq:negative eps},
    when $ \Omega_{j} = \min (\frac{x_{j}}{\epsilon}, \frac{1-x_{j}}{\epsilon})$,
    we have $ 0 \leq x_{j} + \Omega_{j} \cdot \delta_{j} \leq 1$
    for each pixel $j$.
    This indicates that if $ \bm{\Omega} = \min (\frac{ \bm{x} }{\epsilon}, \frac{1- \bm{x}}{\epsilon})$,
    we always have $\bm{x} + \bm{\Omega} \odot \bm{\delta} \in [0, 1]^{\textrm{d}}$.
    In other words, the resulting adversarial example is always valid,
    \ie\ $ \bm{x} ~+~ \bm{w}^{\prime} \odot \bm{\delta} \in [0, 1]^d$.

\end{proof}

\section{Algorithm Flow of Our Approach} 
\label{sec:sup-algorithm}

This section supports the main paper by 
presenting the algorithm flow of our approach
in Algorithm~\ref{alg:wi-fgsm}.
%
% algorithm
% \input{C-2-algorithm.tex}
%
\begin{algorithm} [tb] % [htbp] 
    \caption{Sparse Attack}
    \label{alg:wi-fgsm}
    \footnotesize
 \begin{algorithmic}
    \STATE {\bfseries Input:} natural image $\bm{x}$, 
    the classifer $f_{\bm{\theta}}$, the cross-entropy loss function $J$, 
    adversarial label $y_{\textrm{adv}}$ \\
    \STATE {\bfseries Output:}
    sparse perturbation $\bm{\delta}^{\ast}$
    \STATE \textbf{Initialize:} the weight $\bm{w}$, 
        the $l_{\infty}$ constraint $\epsilon$,
        hyperparameters $\lambda$ and $\tau$
        
        \STATE $\bm{\delta} = $ I-FGSM~$(\epsilon, ~\bm{x}, ~y_{\textrm{adv}})$
        \STATE $\bm{\Omega} = \min (\frac{\bm{x}}{\epsilon}, \frac{1-\bm{x}}{\epsilon})$
        \FOR{$t = 1$ to $T$}
        \STATE // Note that $\pi (\cdot)$ and $H(\cdot)$ respectively represent the ReLU and the Heaviside step function
        \STATE $\bm{w}  = \pi ( \bm{w} - \frac{\tau}{\epsilon} )$
        \STATE $J_{\textrm{adv}} = J(f_{\bm{\theta}}(\bm{x} + \bm{w} \odot \bm{\delta}), ~y_{\textrm{adv}}) + \lambda \sum_{j=1}^{d} H ( w_{j})$ 
        \STATE // $\eta$ denotes the learning rate
        \STATE $\bm{w} =  \bm{w} - \eta \cdot \nabla_{\bm{w}} J_{\textrm{adv}} $    %~~// Update $\bm{w}$ with SGD with momentum
        \ENDFOR
        \STATE $\bm{\delta}^{\ast} = \bm{\Omega} \odot H \left( \pi ( \bm{w} - \frac{\tau}{\epsilon} ) \right) \odot \bm{\delta}$
        \STATE \textbf{return} $\bm{\delta}^{\ast}$
 \end{algorithmic}
 \end{algorithm}

\begin{figure} [!t] %[htbp] 
    \captionsetup[subfigure]{justification=centering}
        
        \centering
        \begin{subfigure}[t]{0.2\textwidth}
            \centering
            \includegraphics[width=\textwidth]{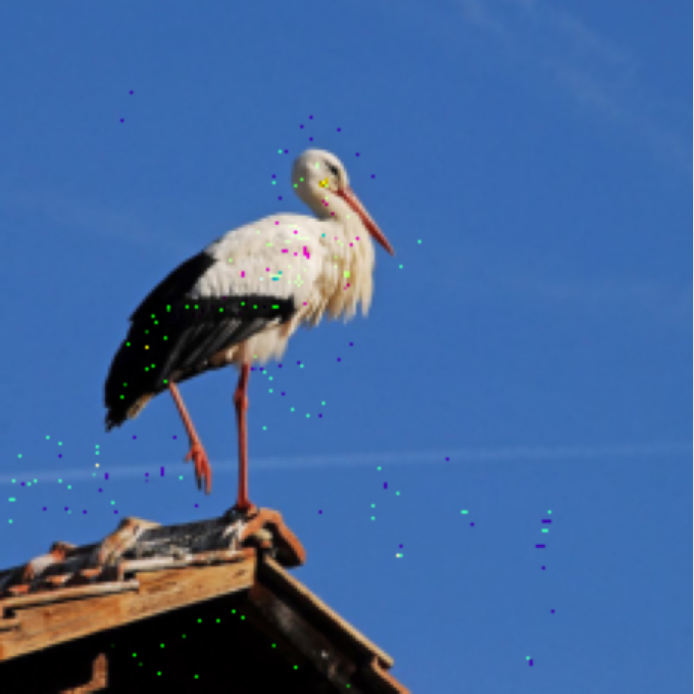}
            \caption{SparseFool. \\ 
                ``pelican" \\
                $ \norm{ \bm{\delta}}_{0} = 248$  }
            \label{fig:exp-visual-sparsefool}
        \end{subfigure}
        \begin{subfigure}[t]{0.2\textwidth}
            \centering
            \includegraphics[width=\textwidth]{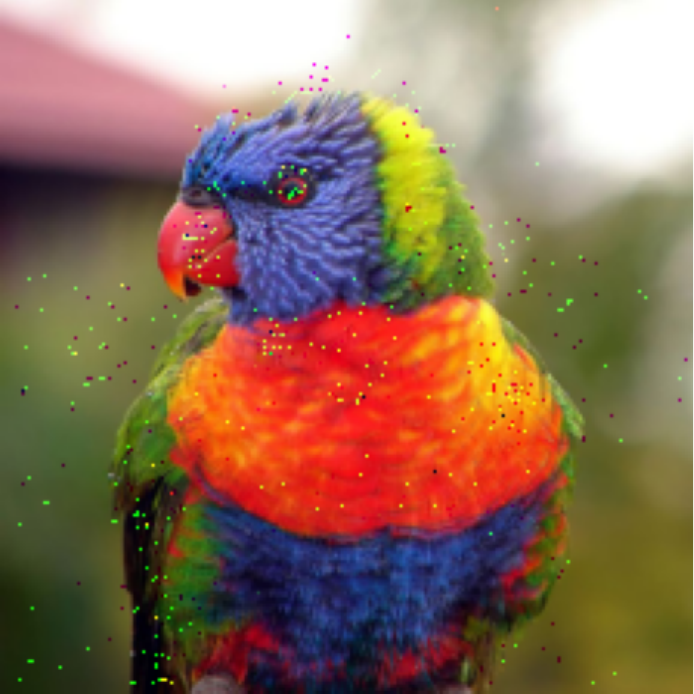}
            \caption{GreedyFool. \\ 
                ``goldfish" \\
                $ \norm{ \bm{\delta}}_{0} = 561$  }
            \label{fig:exp-visual-greedyfool}
        \end{subfigure}
        \begin{subfigure}[t]{0.2\textwidth}
            \centering
            \includegraphics[width=\textwidth]{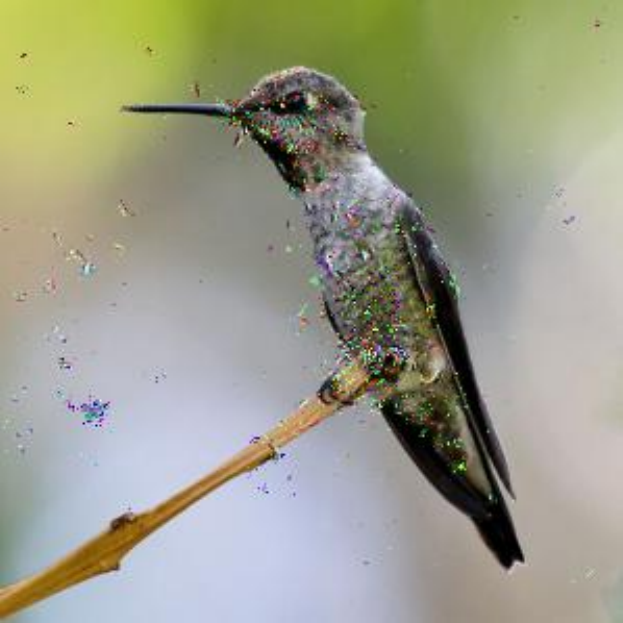}
            \caption{Homotopy. \\ 
                ``stingray" \\
                $ \norm{ \bm{\delta}}_{0} = 4013$ 
            }
            \label{fig:exp-visual-c-ours}
        \end{subfigure}

        \begin{subfigure}[t]{0.2\textwidth}
            \centering
            \includegraphics[width=\textwidth]{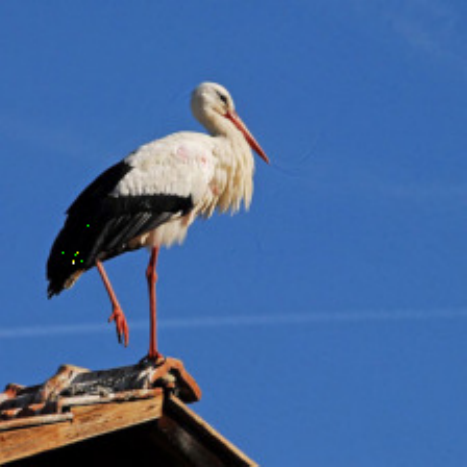}
            \caption{Ours. \\ 
                ``spoonbill" \\
                $ \norm{ \bm{\delta}}_{0} = 22$  }
            \label{fig:exp-visual-a-ours}
        \end{subfigure}
        \begin{subfigure}[t]{0.2\textwidth}
            \centering
            \includegraphics[width=\textwidth]{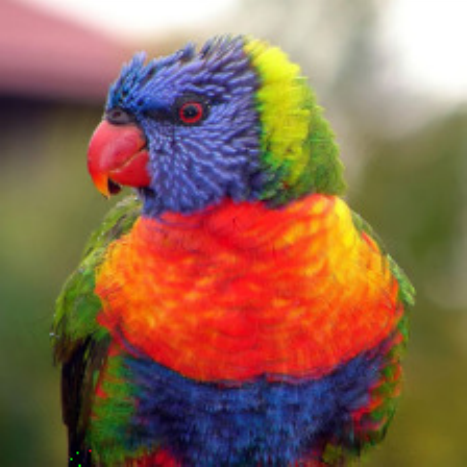}
            \caption{Ours. \\ 
                ``handkerchief" \\
                $ \norm{ \bm{\delta}}_{0} = 10$  }
            \label{fig:exp-visual-b-ours}
        \end{subfigure}
        \begin{subfigure}[t]{0.2\textwidth}
            \centering
            \includegraphics[width=\textwidth]{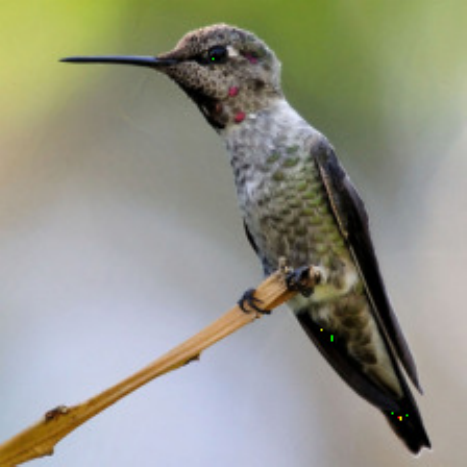}
            \caption{Ours. \\ 
                ``quill" \\
                $ \norm{ \bm{\delta}}_{0} = 26$ 
            }
            \label{fig:exp-visual-c-ours}
        \end{subfigure}
        % \vspace{-0.5em}
        \caption{
            Illustration of adversarial examples computed by our approach and other sparse attacks.
            From left to right, the birds are ``white stork", 
            ``lorikeet", and ``hummingbird", respectively.
            The incorrect prediction and the number of perturbed pixels 
            (\ie\ $\norm{ \bm{\delta}}_{0}$) are listed under each image.
        }
        % \vspace{-2.0 em}
        \label{fig:exp-visual-effect}
\end{figure}

\section{Supporting Experiments} \label{sec:sup-exp}

\subsection{Sparsity under the Non-Targeted Attack}
\label{sec:exp-quality-targeted}

    This section supports Section 5.2 in the main paper by 
    presenting qualitative evaluation on sparsity under the non-targeted attack scenario.
    The experimental settings are the same as Section 5.2 of the main paper,
    except that we perform non-targeted attacks here.
    Specifically, we transform three types of birds from ImageNet into adversarial examples.
    Figure~\ref{fig:exp-visual-effect} shows the qualitative results.
    We observe that our approach achieves the best sparsity, 
    \ie\ modifying only $22, 10, $ and $26$ pixels on ``lorikeet'', ``hummingbird',
    and ``white stork'', respectively, is sufficient to perform effective attacks.
    In contrast, SparseFool~\cite{croce19sparse}, GreedyFool~\cite{dong2020greedyfool}, and Homotopy~\cite{zhu2021homotopy} require 
    perturbing  $248$, $561$, and $4013$ pixels, respectively,  to mislead the same model.
    We also observe that our approach tends to perturb the most relevant contents (\ie\ the bird), 
    which can explain its effectiveness to some extent.

% noise 
\begin{figure}[!t]
    \captionsetup[subfigure]{justification=centering}
    \centering
    \begin{subfigure}[t]{0.2\textwidth}
        \centering
         \includegraphics[width=\textwidth]{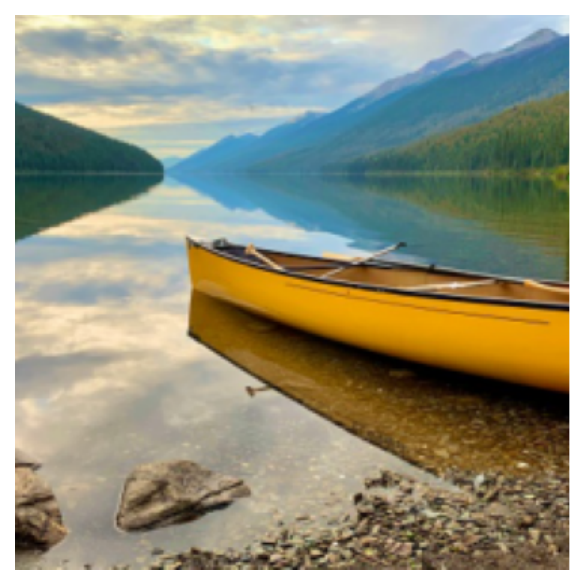}
        \caption{Clean image \\ ``canoe''}
        \label{fig:exp-noise-canoe-clean}
    \end{subfigure}
    \centering
     \begin{subfigure}[t]{0.2\textwidth}
        \centering
         \includegraphics[width=\textwidth]{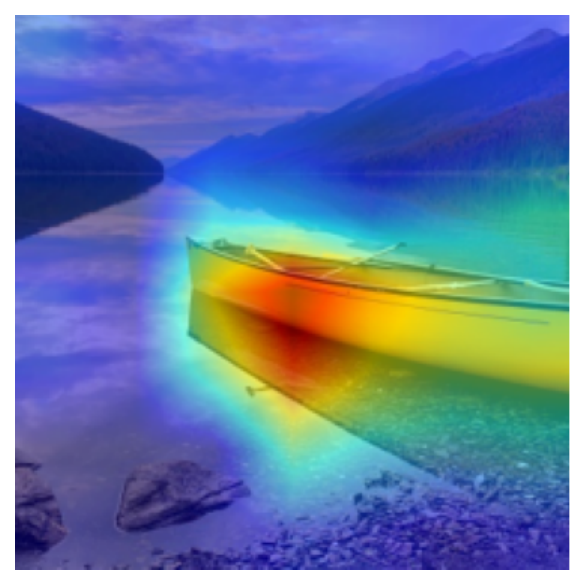}
        \caption{Grad-CAM \\ ``canoe''}
        \label{fig:exp-noise-canoe-grad-cam-clean}
    \end{subfigure}
    \centering
    \begin{subfigure}[t]{0.2\textwidth}
        \centering
         \includegraphics[width=\textwidth]{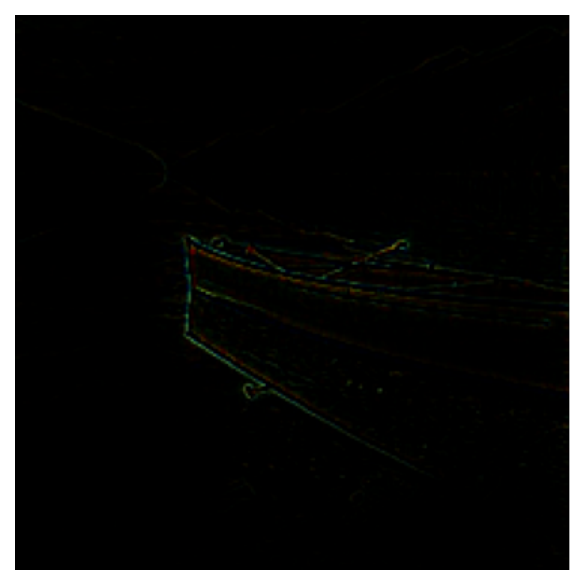}
        \caption{Guided Grad-CAM \\ ``canoe''}
        \label{fig:exp-noise-canoe-guided-grad-cam}
    \end{subfigure}

    \centering
    \begin{subfigure}[t]{0.2\textwidth}
        \centering
         \includegraphics[width=\textwidth]{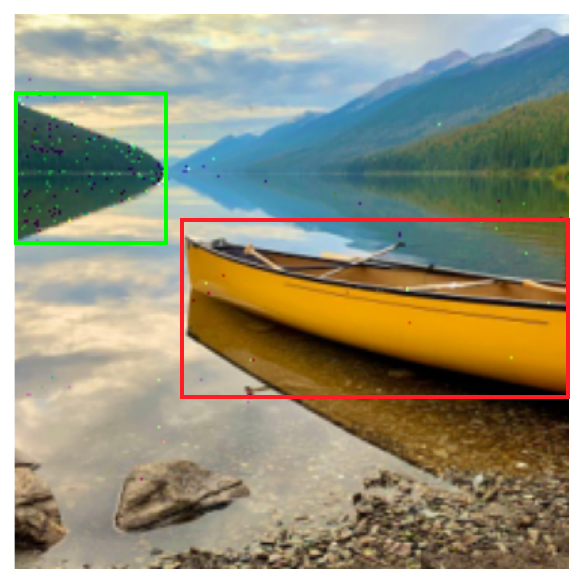}
        \caption{Adversarial example \\ ``wing''}
        \label{fig:exp-noise-canoe-adv}
    \end{subfigure}
     \begin{subfigure}[t]{0.2\textwidth}
        \centering
         \includegraphics[width=\textwidth]{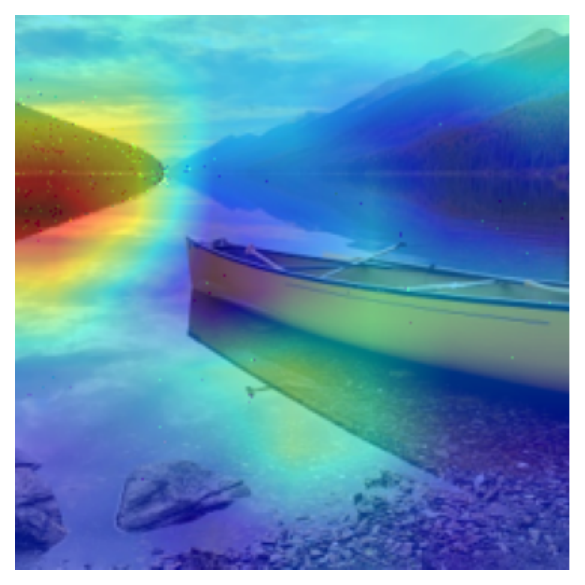}
        \caption{Grad-CAM \\ ``wing''}
        \label{fig:exp-noise-canoe-grad-cam-adv}
    \end{subfigure}
    \begin{subfigure}[t]{0.2\textwidth}
        \centering
         \includegraphics[width=\textwidth]{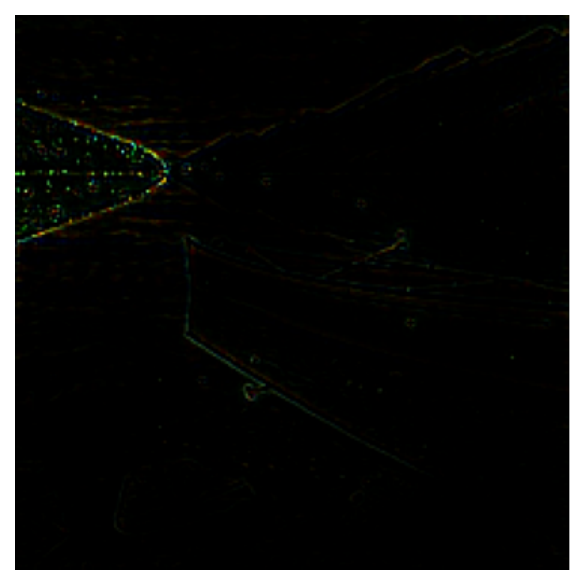}
        \caption{Guided Grad-CAM \\ ``wing''}
        \label{fig:exp-noise-canoe-guided-grad-cam-adv}
    \end{subfigure}
    % \vspace{-0.5em}
    \caption{
        Illustration of how adversarial perturbation computed by our approach
        misleads ResNet-50 to predict an incorrect label of ``wing'' 
        on an image of ``canoe''.
        \textbf{Left:} clean image and adversarial example with their predicted labels.
        \textbf{Middle:} Grad-CAM visualization.
        \textbf{Right:} Guided Grad-CAM visualization.
        }
    % \vspace{-1.5 em}
    \label{fig:exp-noise-canoe}
\end{figure}

\subsection{Interpreting the Adversarial Perturbation}
\label{sec:exp-noise}

This section supports Section 5.3 in the main paper by exhibiting another example of 
how the adversarial perturbation misleads deep neural networks (DNNs) into incorrect predictions.
The experimental settings are the same as the main paper,
except that we perform Grad-CAM and Guided Grad-CAM visualizations~\cite{selvaraju:iccv17:grad-cam} 
on a clean image and its corresponding adversarial example
by using the prediction made by ResNet-50~\cite{he2016deep} as the decision of the internet.
Figure~\ref{fig:exp-noise-canoe} shows how an adversarial example 
produced by our approach misleads ResNet-50 to make an incorrect prediction of ``wing'' on an image of ``canoe''.
We discover that the ``obscuring noise''
(See Figure~\ref{fig:exp-noise-canoe-guided-grad-cam}, perturbations within the red box)
prevents DNNs from identifying the true label by
covering the critical features of ``canoe'',
while the ``leading noise'' 
(See Figure~\ref{fig:exp-noise-canoe-guided-grad-cam}, perturbations within the green box)
misleads DNNs into the targeted prediction
by adding essential features of ``wing''.
These empirical results confirm our discovery in Section 5.3 of the main paper.

\subsection{Ablation Studies}
\label{sup:exp-as}

\begin{figure*}[!t]
    \centering
    \begin{subfigure}[t]{0.29\textwidth}
        \centering
         \includegraphics[width=\textwidth]{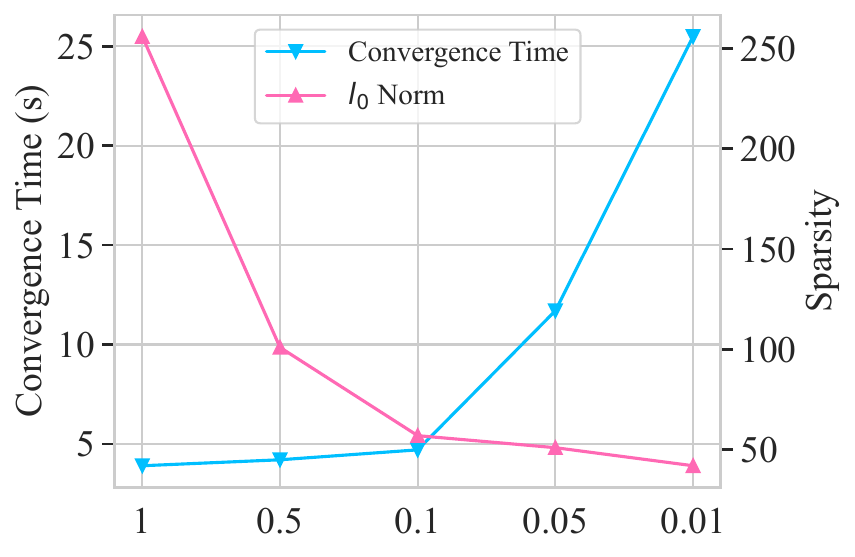}
        %    \vspace{-1em}
        \caption{Hyperparameter $a$}
        \label{fig:exp-sensitivity-a}
    \end{subfigure}
    \centering
    \begin{subfigure}[t]{0.3\textwidth}
        \centering
         \includegraphics[width=\textwidth]{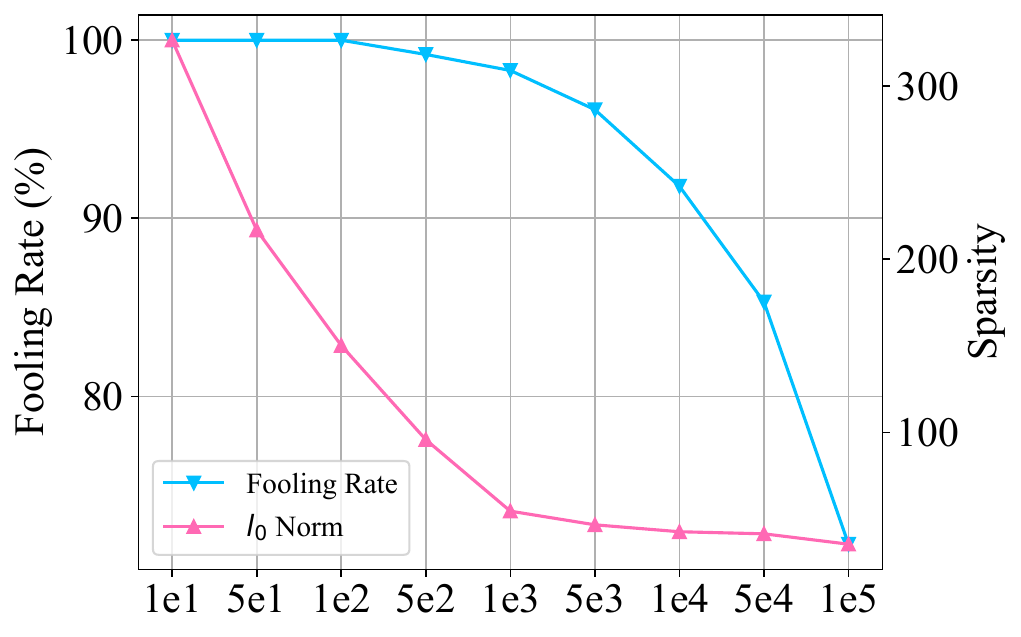}
        %    \vspace{-1em}
        \caption{Hyperparameter $\lambda$}
        \label{fig:exp-sensitivity-lamb}
    \end{subfigure}
    \centering
     \begin{subfigure}[t]{0.3\textwidth}
        \centering
         \includegraphics[width=\textwidth]{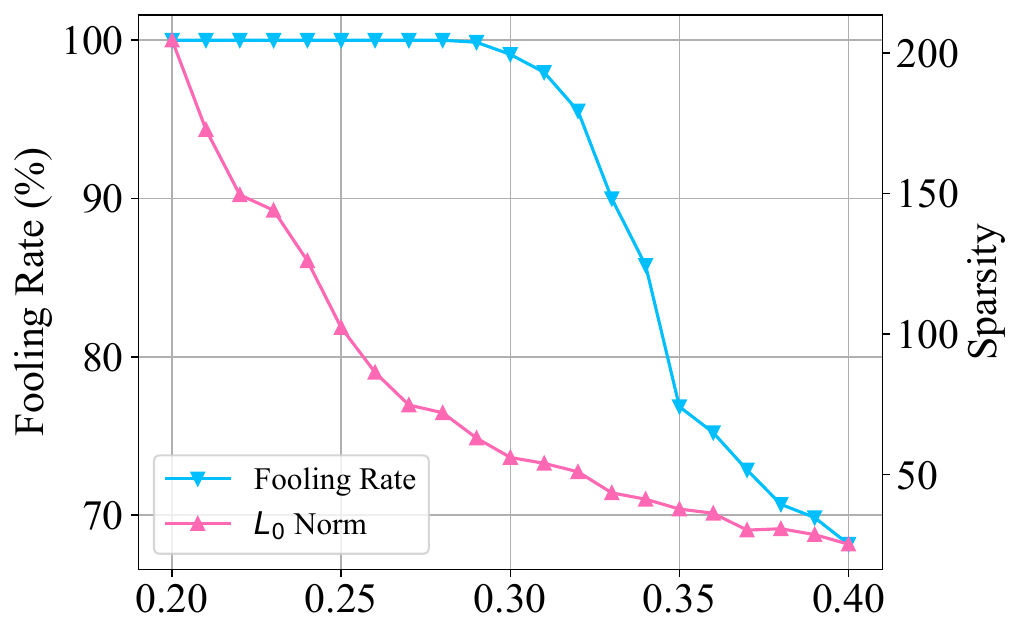}
        %    \vspace{-1em}
        \caption{Hyperparameter $\tau$}
        \label{fig:exp-sensitivity-tau}
    \end{subfigure}
  % \vspace{-0.5em}
    \caption{
        Illustration of how the hyperparameters $a$, $\lambda$ and $\tau$ 
        affect the computational complexity (or fooling rate) and the sparsity.
    }
    \label{fig:exp-sensitivity}
    % \vspace{-1.5 em}
\end{figure*}

We conduct experiments to show the impact of hyperparameters $a$, $\lambda$, and $\tau$ on performance. 
We use VGG-16 to perform white-box attacks under ImageNet.
The constraint $\epsilon$ is set to $6/255$ for all experiments.
First, we gradually reduce the hyperparameter $a$ from $1$ to $0.01$. Figure~\ref{fig:exp-sensitivity-a} illustrates the impact of hyperparameter $a$ on computational complexity and sparsity. We observe that a decrease in $a$ enhances sparsity but compromises computational efficiency. This is attributed to $q_{a}(x)$ becoming more akin to the Dirac delta function as $a$ approaches zero,
making our loss function (\ie\ Eq. (20) of the main paper)
more likely to reach the global optima but very hard to optimize.
An optimal balance was achieved at $a = 0.1$, where our approach reaches convergence in $4.7$s and attained a sparsity level of $57$.

Then, we vary hyperparameter $\lambda$'s values from $1e1$ to $1e5$.
Figure~\ref{fig:exp-sensitivity-lamb} illustrates the fooling rate and the sparsity with the variation of  $\lambda$. 
We observe that the hyperparameter $\lambda$ positively affects
the sparsity but negatively influences the fooling rate.
In particular, when $\lambda=1e3$, 
our approach achieves the best tradeoff,
with the fooling rate of $98.3 \%$ and the perturbed pixels of $54$ on average.

Finally, we use grid search to tune the hyperparameter $\tau$
by varying $\tau$ from $0.2$ to $0.4$ in an evenly spaced interval of $0.01$.
Figure~\ref{fig:exp-sensitivity-tau} depicts the effect of $\tau$ on
the fooling rate and the sparsity. 
We observe that an increase in $\tau$ benefits the sparsity but hurts the fooling rate.
The best tradeoff is achieved at  $\tau=0.30$, with $99.1 \%$ fooling rate and  $56$ perturbed pixels on average.

\end{document}